\documentclass[acmsmall]{acmart} 
\AtBeginDocument{%
  \providecommand\BibTeX{{%
    \normalfont B\kern-0.5em{\scshape i\kern-0.25em b}\kern-0.8em\TeX}}}

\acmISBN{978-1-4503-XXXX-X/18/06}

\usepackage{amsmath} 

\usepackage{graphicx}
\usepackage{subcaption}
\usepackage{threeparttable}
\usepackage{amsmath}
\usepackage{dsfont}
\usepackage[linesnumbered,ruled,vlined]{algorithm2e}
\usepackage{enumitem}

\usepackage{amsmath,amsfonts,mathrsfs,amsthm,stmaryrd}

\usepackage{multirow}

\makeatletter

\@ifundefined{@printauthorsaddresses}{}{
  \renewcommand\@printauthorsaddresses{}
}
\makeatother

\begin{document}
\graphicspath{{../pdf/}{../jpeg/}}
\DeclareGraphicsExtensions{.pdf,.jpeg,.png}

\renewcommand{\eqref}[1]{Eq.\,(\ref{#1})}

\hyphenation{op-tical net-works semi-conduc-tor}

\newcommand{\red}[1]{\textcolor{red}{\textbf{#1}}}
\newcommand{\blue}[1]{\textcolor{blue}{#1}}

\renewcommand{\eqref}[1]{Eq.\,(\ref{#1})}

\newcommand*\diff{\mathop{}\!\mathrm{d}}
\renewcommand{\Pr}{\textnormal{P}}

\renewcommand{\vec}[1]{\mathbf{#1}}
\newcommand{\x}[1]{\mathbf{x}_{#1}}
\newcommand{\z}[1]{\mathbf{z}_{#1}}
\newcommand{\f}{\mathbf{f}}
\newcommand{\T}{^{\top}}
\newcommand{\xfull}{\x{t}^{\textnormal{rec}}}

\newcommand{\U}[1]{\mathcal{U}_{#1}}

\newcommand{\xold}[1]{\mathbf{x}_{#1}^{share}}
\newcommand{\xnew}[1]{\mathbf{x}_{#1}^{new}}
\newcommand{\xre}[1]{\tilde{ \mathbf{x}_{#1} }}
\newcommand{\xT}[1]{\mathbf{x}_{#1}^{\mkern-5mu\top}}
\newcommand{\xR}[1]{\tilde{\mathbf{x}}_{#1}}
\newcommand{\w}[1]{\mathbf{w}_{#1}}
\newcommand{\wT}[1]{\mathbf{w}_{#1}^{\mkern-5mu\top}}
\newcommand{\wA}[1]{\bar{\mathbf{w}}_{#1}}
\newcommand{\wR}[1]{\tilde{\mathbf{w}}_{#1}}
\newcommand{\wAT}[1]{\bar{\mathbf{w}}_{#1}^{\mkern-0mu\top}}
\newcommand{\wRT}[1]{\hat{\mathbf{w}}_{#1}^{\mkern-0mu\top}}
\newcommand{\yA}[1]{\hat{y}_{\text{obs}}}
\newcommand{\yR}[1]{\hat{y}_{\text{rec}}}
\newcommand{\LO}[1]{L_{#1}^{\text{obs}}}
\newcommand{\LR}[1]{L_{#1}^{\text{rec}}}
\newcommand{\LE}[1]{L_{#1}}
\renewcommand{\u}[1]{\mathbf{u}_{#1}}
\newcommand{\GR}{\mathbf{G\mkern-1mu r}}
\newcommand{\IT}{\mathbf{I}_{t}}
\newcommand{\romup}[1]{\uppercase\expandafter{\romannumeral #1\relax}}
\newcommand{\romlo}[1]{\lowercase\expandafter{\romannumeral #1\relax}}

\newcommand*{\argIn}{\makebox[1ex]{\textbf{$\cdot$}}}
\newcommand{\M}[1]{\mathbf{M}_{#1}}
\newcommand{\Rset}[1]{\mathbb{R}^{#1}}
\newcommand{\ts}{\textsuperscript}
\newcommand{\pl}{\ell}

\newcommand{\ie}{\textit{i}.\textit{e}., }
\newcommand{\eg}{\textit{e}.\textit{g}., }
\newcommand{\cf}{\textit{cf}.}
\newcommand{\wrt}{\textit{w}.\textit{r}.\textit{t}.}
\newcommand{\etc}{\textit{etc}.}
\newcommand{\rom}[1]{\uppercase\expandafter{\romannumeral #1\relax}}
\newcommand{\romsm}[1]{\lowercase\expandafter{\romannumeral #1\relax}}

\renewcommand{\thefootnote}{\roman{footnote}}
\newcommand{\prd}{OLVFS}
\newcommand{\VS}{capricious data streams}

\renewcommand{\eqref}[1]{Eq.\,(\ref{#1})}
\renewcommand{\Pr}{\textnormal{P}}
\newtheorem{RQs}{RQs}
\newtheorem{remark}{Remark}

\renewcommand{\vec}[1]{\mathbf{#1}}
\renewcommand{\u}[1]{\mathbf{u}_{#1}}
\newcommand{\alg}{OL-MDISF}
\newcommand{\Capr}{capricious data streams}
\newcommand{\Trap}{Trapezoidal data streams}
\newcommand{\SSL}{$OS^{2}L$}

\title{Extension OL-MDISF: Online Learning from Mix-Typed, Drifted, and Incomplete Streaming Features}

\author{Shengda Zhuo}
\orcid{0000-0001-5610-005X}
\affiliation{
  \institution{College of Cyber Security, Jinan University, Guangzhou, Guangdong}
  \postcode{510632}
  \country{China}
}
\email{zhuosd96@gmail.com}

\author{Di~Wu}
\orcid{0000-0002-7788-9202}
\authornote{The Correspondence author.}
\affiliation{
  \institution{College of Computer and Information Science, Southwest University, Chongqing}
  \postcode{400715}
  \country{China}
}

\email{wudi.cigit@gmail.com}
\author{Yi~He}
\orcid{0000-0002-5357-6623}
\affiliation{
  \institution{School of Computing, Data Science, and Physics, William \& Mary, Williamsburg, VA}
  \postcode{23185}
  \country{USA}
}
\email{yihe@wm.edu}

\author{Shuqiang Huang}
\orcid{0000-0001-9551-022X}
\affiliation{
  \institution{College of Cyber Security, Jinan University, Guangzhou, Guangdong}
  \postcode{510632}
  \country{China}
}
\email{hsq@jnu.edu.cn}

\author{Xindong Wu}
\orcid{0000-0003-2396-1704}
\affiliation{
  \institution{Key Laboratory of Knowledge Engineering with Big Data (the Ministry of Education of China), School of Computer Science and Information Engineering, Hefei University of Technology, Hefei, Anhui}
  \postcode{230009}
  \country{China}
}
\email{xwu@hfut.edu.cn}


\begin{abstract}

Online learning, where feature spaces can change over time, offers a flexible learning paradigm that has attracted considerable attention.
However, it still faces three significant challenges. 
First, the heterogeneity of real-world data streams with mixed feature types presents challenges for traditional parametric modeling. 
Second, data stream distributions can shift over time, causing an abrupt and substantial decline in model performance.
Additionally, the time and cost constraints make it infeasible to label every data instance in a supervised setting.
To overcome these challenges, we propose a new algorithm \emph{Online Learning from Mix-typed, Drifted, and Incomplete Streaming Features} (\alg), which aims to relax restrictions on both feature types, data distribution, and supervision information.
Our approach involves utilizing copula models to create a comprehensive latent space, employing an adaptive sliding window for detecting drift points to ensure model stability, and establishing label proximity information based on geometric structural relationships. 
To demonstrate the model's efficiency and effectiveness, we provide theoretical analysis and comprehensive experimental results.

This extension serves as a standalone technical reference to the original OL-MDISF\footnote{https://dl.acm.org/doi/10.1145/3744712} method. It provides (i) a contextual analysis of OL-MDISF within the broader landscape of online learning, covering recent advances in mixed-type feature modeling, concept drift adaptation, and weak supervision, and (ii) a comprehensive set of experiments across 14 real-world datasets under two types of drift scenarios. These include full CER trends, ablation studies, sensitivity analyses, and temporal ensemble dynamics. We hope this document can serve as a reproducible benchmark and technical resource for researchers working on nonstationary, heterogeneous, and weakly supervised data streams.

\end{abstract}


\maketitle

\section{Introduction}

Online learning in nonstationary environments has become increasingly important in modern data-driven applications such as real-time monitoring, personalized recommendation, financial risk modeling, and autonomous decision systems. These scenarios often involve data streams that continuously evolve over time, bringing about new information patterns and behaviors. Unlike traditional offline learning paradigms, online models must adapt quickly to incoming instances, operate under limited supervision, and be robust to uncertainties in both input space and label space.

To meet these demands, a wide range of online learning methods have been proposed. Some focus on adaptation to concept drift via dynamic windows or model ensembles, while others study learning from evolving feature spaces or perform online feature selection. In parallel, semi-supervised or active learning techniques have been explored to handle sparse or delayed labels. Despite these advances, most methods concentrate on individual challenges in isolation—e.g., either drift, or label sparsity, or feature evolution—without providing a unified treatment of all three.

In practice, data streams are often \emph{mix-typed} (containing numerical, categorical, and missing values), \emph{nonstationary} (subject to evolving distributions and abrupt concept shifts), and \emph{incompletely labeled} (only a fraction of instances are annotated). These properties interact in complex ways and pose major obstacles for traditional parametric models. For example, missing values and inconsistent feature schemas hinder alignment across time, while irregular labels weaken the reliability of feedback for model updates. Building a unified online learner that simultaneously handles heterogeneity, drift, and supervision scarcity remains a largely unsolved challenge.

To address these challenges, our prior work published in ACM Transactions on Knowledge Discovery from Data (TKDD 2025) introduced \textbf{OL-MDISF} (Online Learning from Mix-typed, Drifted, and Incomplete Streaming Features)~\cite{zhuo2025online}, a copula-based learning framework that:
\begin{itemize}[leftmargin=*]
  \item constructs latent representations across mix-typed features using copula models,
  \item detects and responds to distributional changes via adaptive ensemble-based drift tracking,
  \item and performs structure-aware pseudo-labeling in geometric latent space to mitigate sparse supervision.
\end{itemize}
OL-MDISF integrates these components into a unified, online, end-to-end model capable of operating under extreme streaming conditions.

While the TKDD article introduced OL-MDISF’s algorithmic foundations and presented high-level experimental validation, many important empirical details were deferred due to space constraints. This supplementary manuscript provides a \textbf{comprehensive yet standalone extension}, including:
\begin{itemize}[leftmargin=*]
  \item an in-depth contextual discussion of related work in mix-typed online learning, drift adaptation, and weak supervision,
  \item complete experiments on 14 real-world datasets under two distinct drift settings (capricious and trapezoidal),
  \item ablation studies isolating the impact of individual model components,
  \item sensitivity analyses with respect to varying label missingness, and temporal analysis of ensemble dynamics across streaming steps.
\end{itemize}
Our goal is to offer the research community a reproducible and self-contained technical reference that deepens the understanding of OL-MDISF and supports future developments in streaming machine learning.

\section{Background and Related Work}
\label{sec:theory}
The online learning paradigm has garnered significant attention due to its capability to adapt to evolving data in real-time scenarios. However, three major challenges persist in real-world streaming environments: (i) heterogeneous feature types, (ii) nonstationary distributions (concept drift), and (iii) incomplete supervision. OL-MDISF was designed to address all three simultaneously, whereas most prior work focuses on subsets of these challenges.

\subsection{Mixed-Type and Evolving Feature Spaces}
In real-world streaming environments, the feature space is often dynamic—new features may appear over time, existing features may vanish or shift, and the data types may vary across instances. These \emph{mix-typed evolving streams} introduce a dual challenge: managing heterogeneity in data types (\emph{e.g.,} categorical, numerical, and missing) and coping with temporal inconsistency in feature dimensionality.

A prominent line of work tackles feature space evolution via online feature selection (OFS). These approaches iteratively select a sparse set of relevant features over time to improve scalability and adaptability. Early frameworks such as Conditional Independence OFS~\cite{you2018online}, Capricious Streaming OFS~\cite{wu2019online}, and OSFS-Vague~\cite{yang2024osfs} emphasized dynamic filtering under distributional uncertainty. Later works introduced causal reasoning~\cite{zhang2024fairness}, vague sets~\cite{yang2024osfs}, and graph structures~\cite{chen2025federated} to enhance interpretability and cross-domain adaptation.

Another stream of research seeks to learn latent representations from high-dimensional or irregular streaming data. Techniques such as robust sparse modeling~\cite{chen2024robust}, variational embedding~\cite{wu2023mmlf}, and uncertainty-aware encoding~\cite{wu2023robust} help mitigate the impact of feature type heterogeneity and missingness. These methods often rely on generative assumptions or assume certain stability in the data schema, limiting their applicability to capricious or mix-typed input.

A third family of solutions leverages online matrix decomposition, such as CUR or PCA variants, to dynamically approximate the evolving feature subspace. Works such as online CUR~\cite{qiu2025online,chen2025l1}, lifelong decomposition~\cite{he2021unsupervised}, and norm-regularized adaptation~\cite{chen2024l} have demonstrated empirical success in stream scenarios, especially when the feature arrival pattern is gradual or partially observable.

\par\smallskip\noindent
\textbf{Limitations and Motivation for OL-MDISF.}
Despite these advances, most prior approaches assume a relatively clean or consistent feature schema, and few directly address the joint problem of feature heterogeneity \emph{and} structural evolution. In contrast, OL-MDISF introduces a novel \emph{copula-based latent modeling} strategy, which explicitly models dependencies across mixed-type variables and seamlessly integrates new features into the latent space without requiring imputation or type normalization. This flexible representation allows OL-MDISF to adapt to highly dynamic, noisy, and heterogeneous data streams with minimal assumptions.



\subsection{Concept Drift and Streaming Adaptation}


In nonstationary environments, the data distribution often changes over time, a phenomenon known as \emph{concept drift}. This may arise from seasonality, user behavior change, system evolution, or adversarial shifts. To maintain predictive performance in such settings, online models must be capable of detecting distributional changes and updating themselves accordingly.

A large body of research has explored adaptive mechanisms to cope with drift. Traditional solutions include sliding windows, decay factors, and change-point detection algorithms. For example, ensemble-based techniques~\cite{he2023towards,schreckenberger2023online} dynamically weight or replace weak learners over time to preserve diversity and robustness. Some approaches utilize explicit drift detectors that monitor error rates or confidence intervals to trigger model resets or updates.

More recent studies have adopted probabilistic frameworks or latent variable modeling to capture gradual or hidden changes. Probabilistic drift detectors~\cite{lian2024utilitarian,tang2023online,he2019online} estimate distributional distances over time using likelihood ratios, kernel embeddings, or Bayesian inference. Others extend this view with structural modeling of data shifts across both feature and label spaces, enabling joint drift reasoning and adaptive supervision~\cite{he2021online,lian2022online,beyazit2020online}.

The challenge of drift has further motivated research into related domains such as fairness-aware adaptation~\cite{zhang2024fairness}, federated drift-tolerant learning~\cite{chen2025federated}, and utilitarian learning~\cite{lian2024utilitarian} in open-world scenarios. These directions expand the notion of concept drift to include population shift, group-level fairness violations, or long-term utility shifts.

\par\smallskip\noindent
\textbf{OL-MDISF Contribution.}
OL-MDISF contributes to this literature by incorporating a \emph{temporal drift detector} based on two complementary signals: (i) \emph{ensemble entropy}, which reflects internal disagreement across component models, and (ii) \emph{latent mismatch}, which tracks deviation in the learned copula-based representation. This dual mechanism enables robust, real-time detection of concept drift without relying on ground-truth labels, and supports smooth model transitions even under highly capricious data streams. By coupling this mechanism with a dynamic ensemble strategy, OL-MDISF achieves strong temporal adaptability while avoiding catastrophic forgetting.

\subsection{Learning with Incomplete or Noisy Labels}

In real-time or streaming environments, labeled data is often sparse, delayed, or noisy due to cost, latency, or privacy constraints. This \emph{incomplete supervision} problem significantly hinders model reliability, especially in early learning stages or under sudden distribution shifts. Traditional supervised learners require a steady flow of labeled instances to maintain performance—an assumption rarely satisfied in streaming scenarios.

To mitigate label scarcity, one common line of research is online \emph{active learning}, which selectively queries informative instances from the stream based on uncertainty or diversity~\cite{he2020active,chen2024l}. Other works adopt \emph{semi-supervised learning} schemes~\cite{wu2023online}, where a small labeled subset is used to propagate supervision to unlabeled data using consistency regularization, clustering, or graph-based heuristics.

More recent methods explore \emph{structure-aware learning}, leveraging temporal or geometric dependencies in the data stream to impute labels or improve robustness. Examples include geometry-guided imputation~\cite{you2023online,you2024online}, latent structure modeling under uncertainty~\cite{chen2024robust}, and decision-risk–aware feature selection~\cite{xu2023online}. These approaches show promise but often rely on specific assumptions such as smooth label manifolds or static feature distributions.

Another line of work focuses on modeling label noise and prediction uncertainty explicitly~\cite{beyazit2020online,chen2022online}. These methods typically incorporate confidence estimates, noise correction layers, or Bayesian reasoning into the learning process. While effective in batch settings, their extension to dynamic, mix-typed, or highly drifting streams remains challenging.

\paragraph{OL-MDISF Contribution.}
OL-MDISF addresses incomplete supervision by operating in a \emph{copula-induced latent space}, where the geometry among instances is preserved across heterogeneous input types. It propagates pseudo-labels based on local geometric proximity, allowing supervision signals to spread across both labeled and unlabeled regions of the stream. This structure-aware propagation is inherently uncertainty-tolerant and does not rely on external label query mechanisms. Furthermore, OL-MDISF's latent drift detector supports label-aware updates even in the absence of ground truth, providing robustness to both missingness and noise in evolving environments.

\par\noindent
\textbf{Summary:}
OL-MDISF presents a unified framework that addresses evolving feature representations, concept drift, and incomplete labels within one model. The works cited above tackle these problems individually or partially; our approach offers a holistic solution, and this supplementary study further evaluates its robustness and generalizability under real-world stream settings.


\section{Additional experiments for \alg}
\label{sec:add}

Section~\ref{sec:Section2} presents the complete experimental results that supplement Figs. 4
and 5
in the main content.

\subsection{Additional Experiments} 
\label{sec:Section2}

\begin{figure*}[!t]
	\centering
	\begin{subfigure}[t]{0.25\linewidth}
		\includegraphics[width=\textwidth]{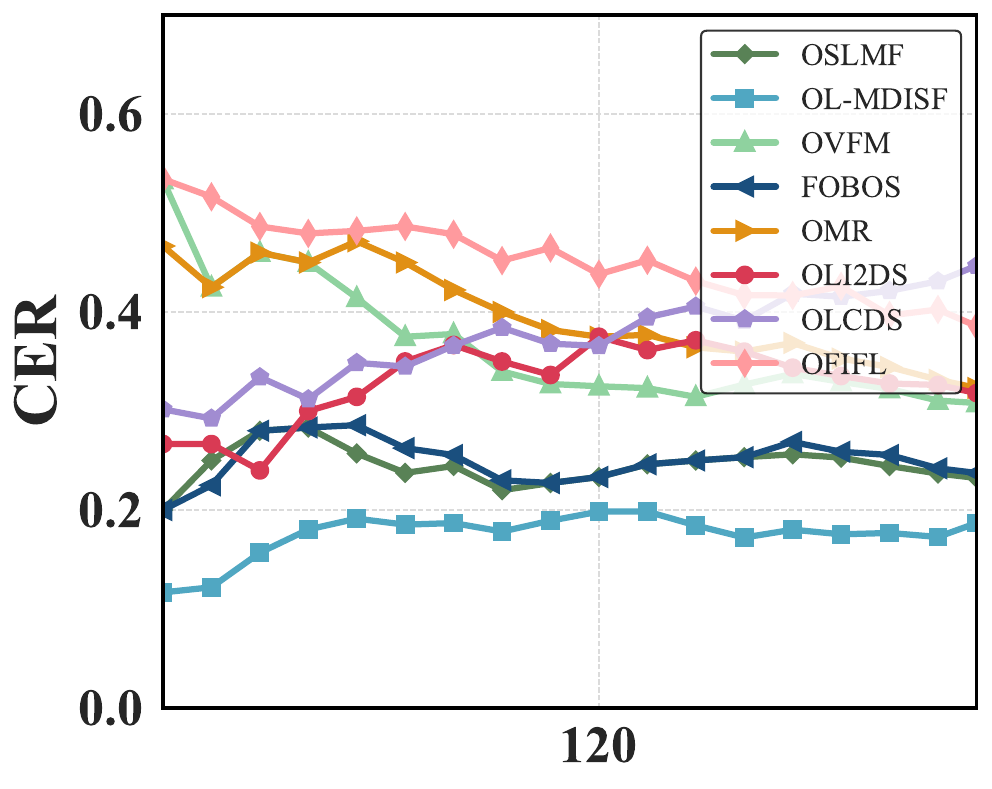}
		\caption{wpbc}
		\label{fig:Carp_wpbc_4_CER}
	\end{subfigure}
	\hspace{2em}
	\begin{subfigure}[t]{0.25\linewidth}
		\includegraphics[width=\textwidth]{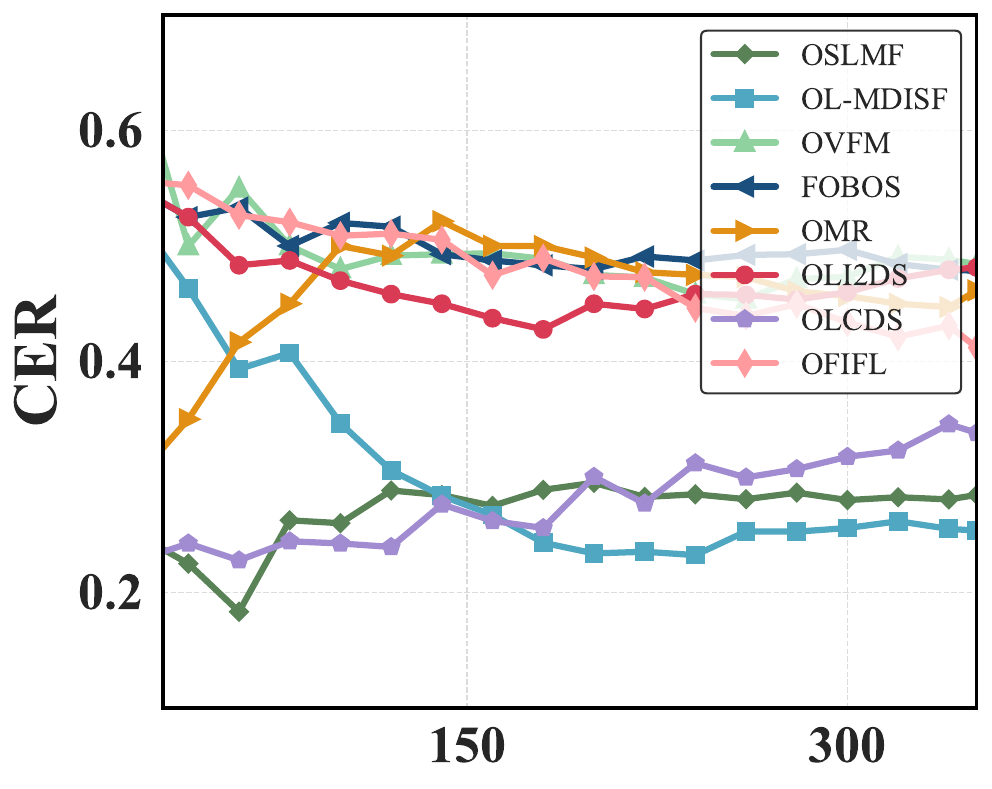}
		\caption{ionosphere}
		\label{fig:Carp_ionosphere_4_CER}
	\end{subfigure}
	\hspace{2em}
	\begin{subfigure}[t]{0.25\linewidth}
		\includegraphics[width=\textwidth]{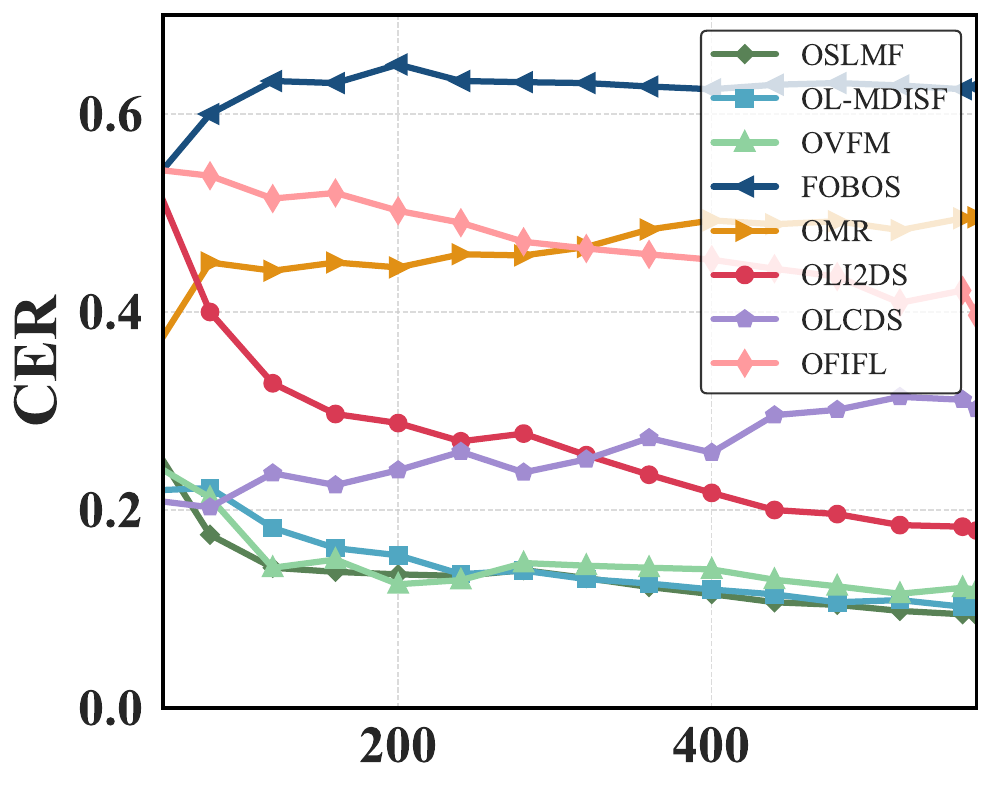}
		\caption{wdbc}
		\label{fig:Carp_wdbc_4_CER}
        \vspace{0.5em}
	\end{subfigure}
	\begin{subfigure}[t]{0.25\linewidth}
		\includegraphics[width=\textwidth]{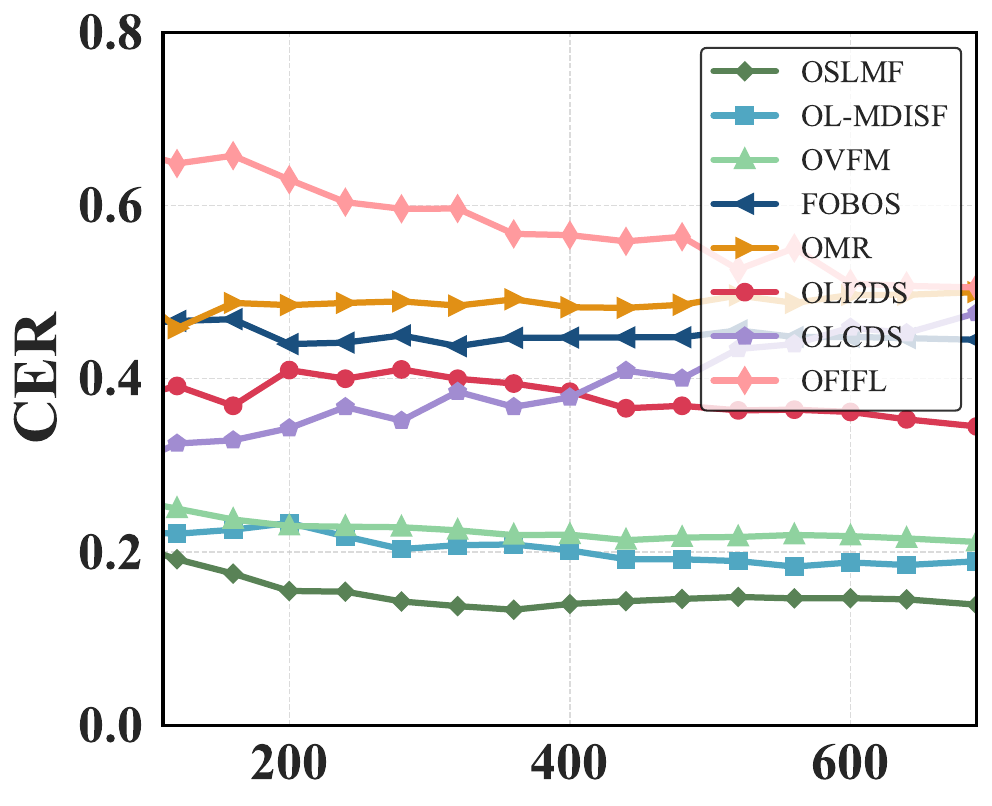}
		\caption{australian}
		\label{fig:Carp_australian_4_CER}
	\end{subfigure}
	\hspace{2em}
	\begin{subfigure}[t]{0.25\linewidth}
		\includegraphics[width=\textwidth]{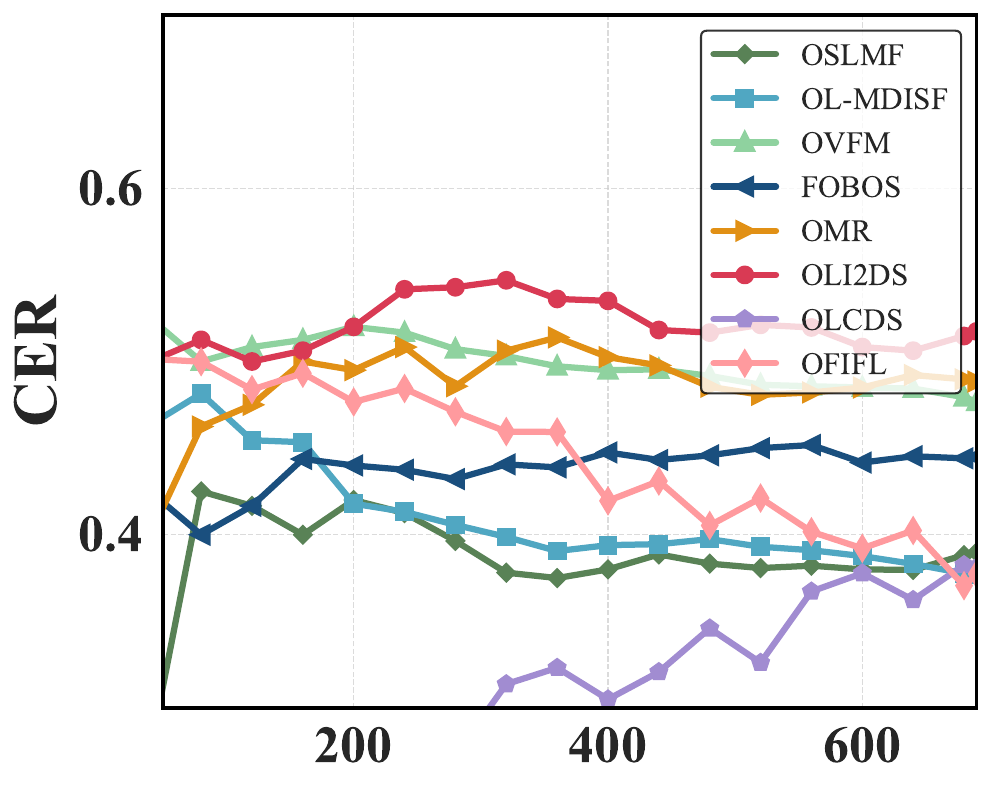}
		\caption{credit-a}
		\label{fig:Carp_credit_4_CER}
	\end{subfigure}
	\hspace{2em}
	\begin{subfigure}[t]{0.25\linewidth}
		\includegraphics[width=\textwidth]{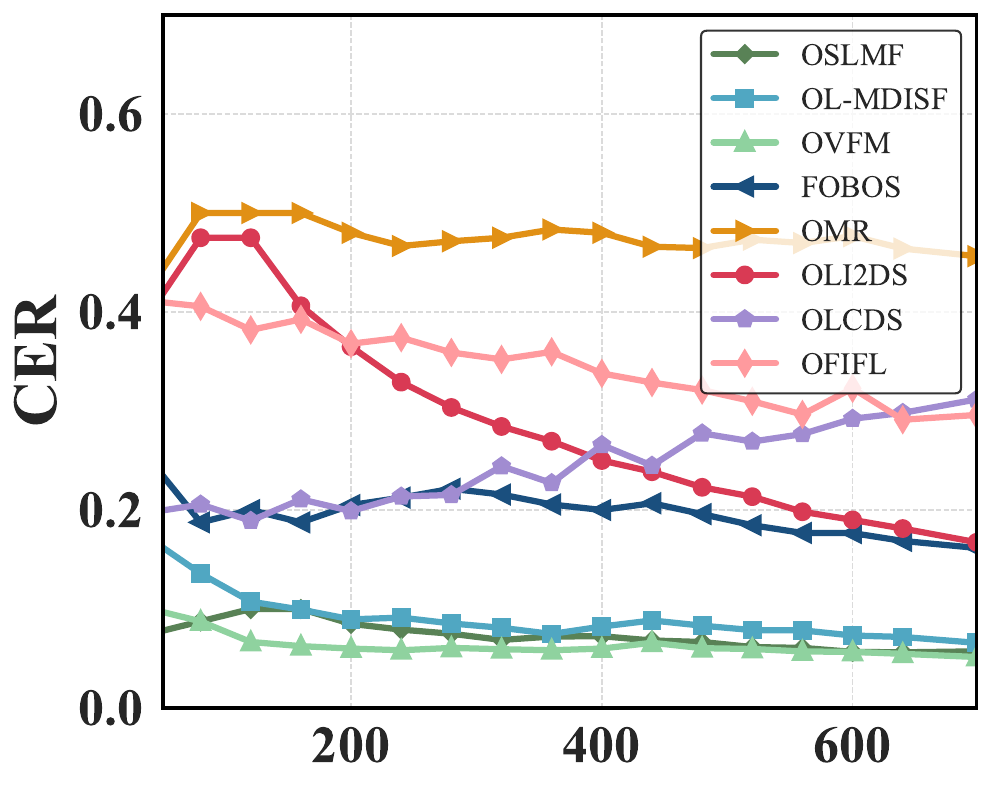}
		\caption{wbc}
		\label{fig:Carp_wbc_4_CER}
        \vspace{0.5em}
	\end{subfigure}
	\begin{subfigure}[t]{0.25\linewidth}
		\includegraphics[width=\textwidth]{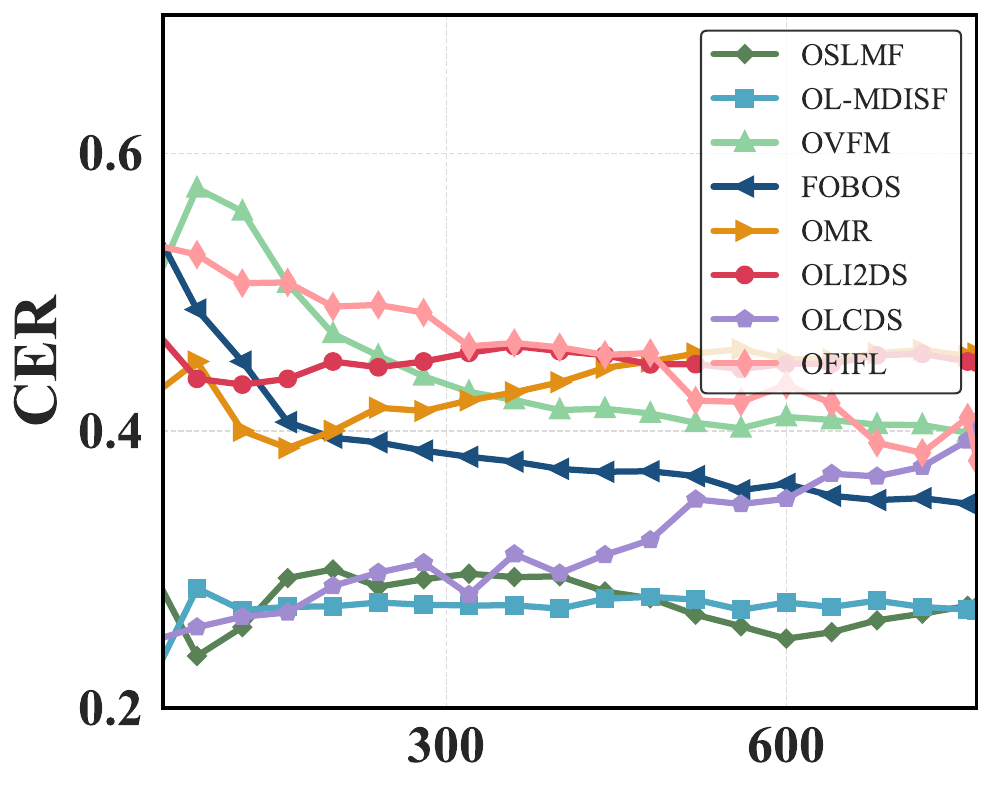}
		\caption{diabetes}
		\label{fig:Carp_diabetes_4_CER}
	\end{subfigure}
	\hspace{2em}
	\begin{subfigure}[t]{0.25\linewidth}
		\includegraphics[width=\textwidth]{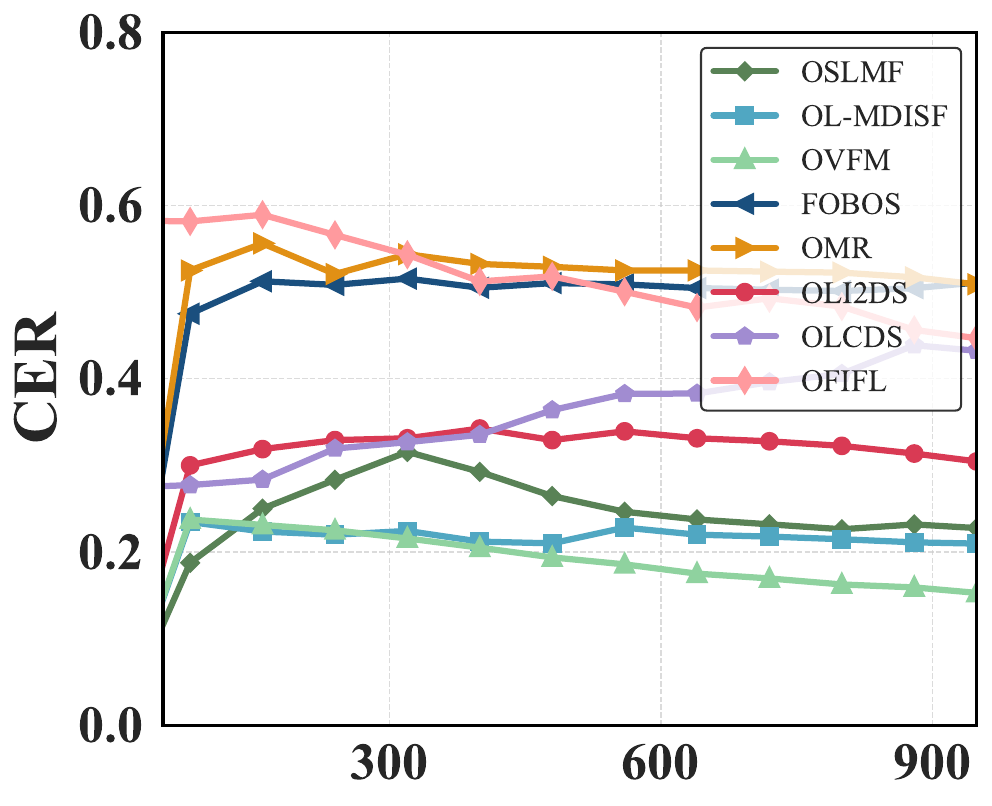}
		\caption{dna}
		\label{fig:Carp_dna_4_CER}
	\end{subfigure}
	\hspace{2em}
	\begin{subfigure}[t]{0.25\linewidth}
		\includegraphics[width=\textwidth]{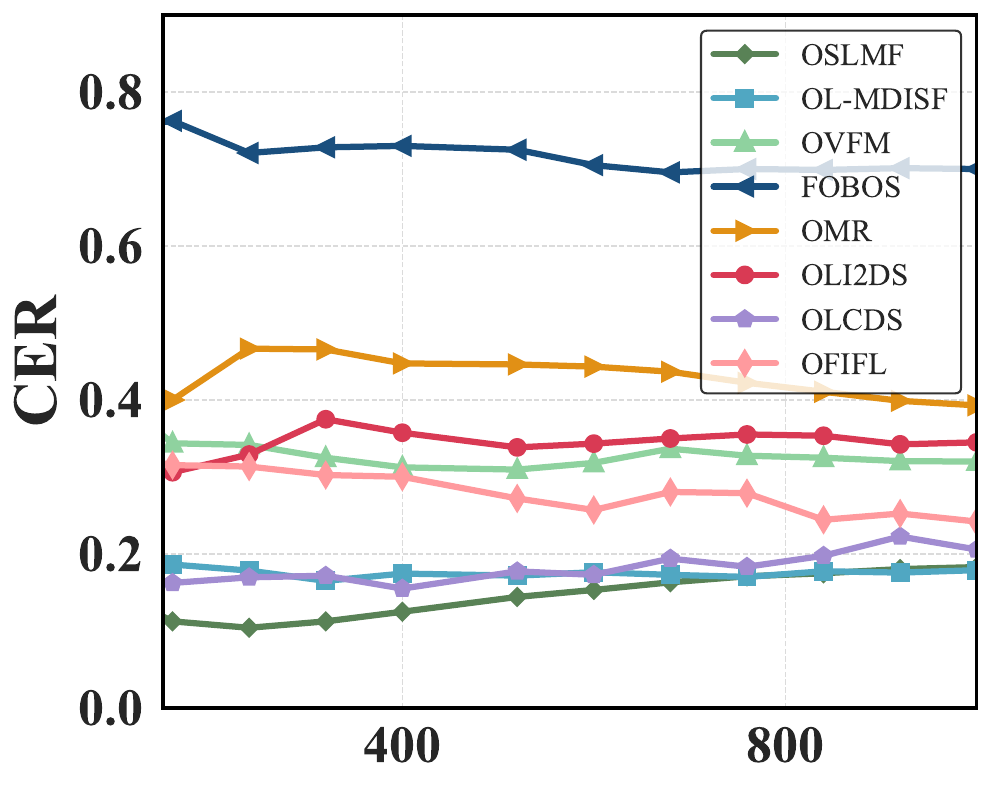}
		\caption{german}
		\label{fig:Carp_german_4_CER}
        \vspace{0.5em}
	\end{subfigure}
	\begin{subfigure}[t]{0.25\linewidth}
		\includegraphics[width=\textwidth]{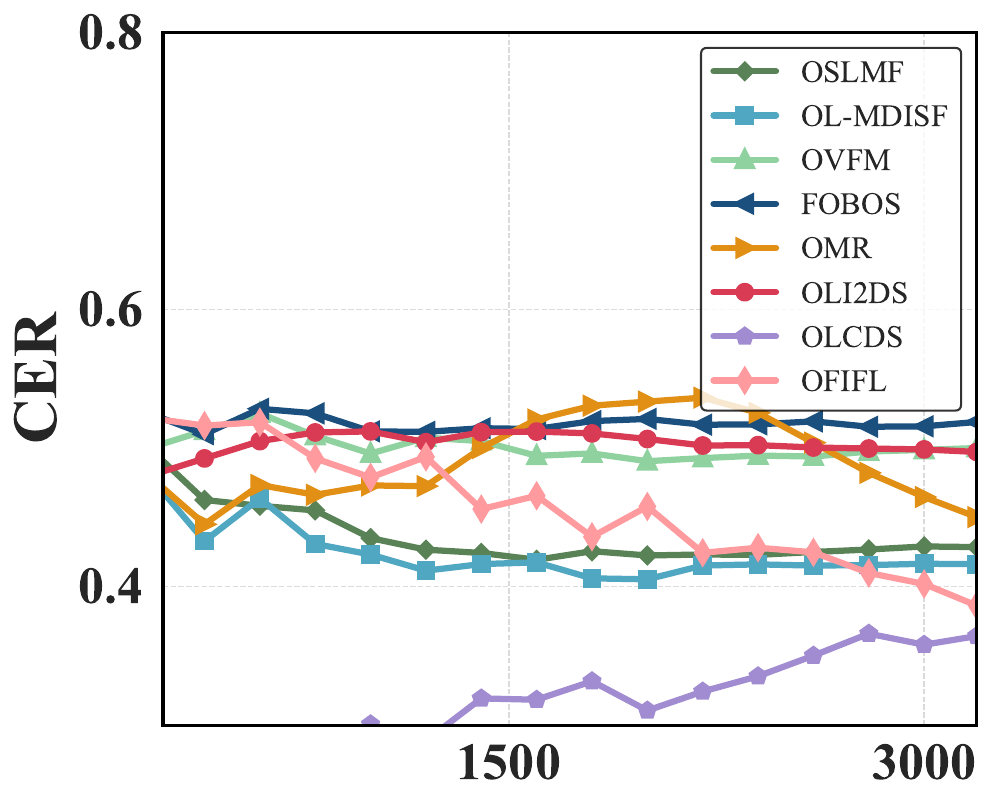}
		\caption{splice}
		\label{fig:Carp_splice_4_CER}
	\end{subfigure}
	\hspace{2em}
	\begin{subfigure}[t]{0.25\linewidth}
		\includegraphics[width=\textwidth]{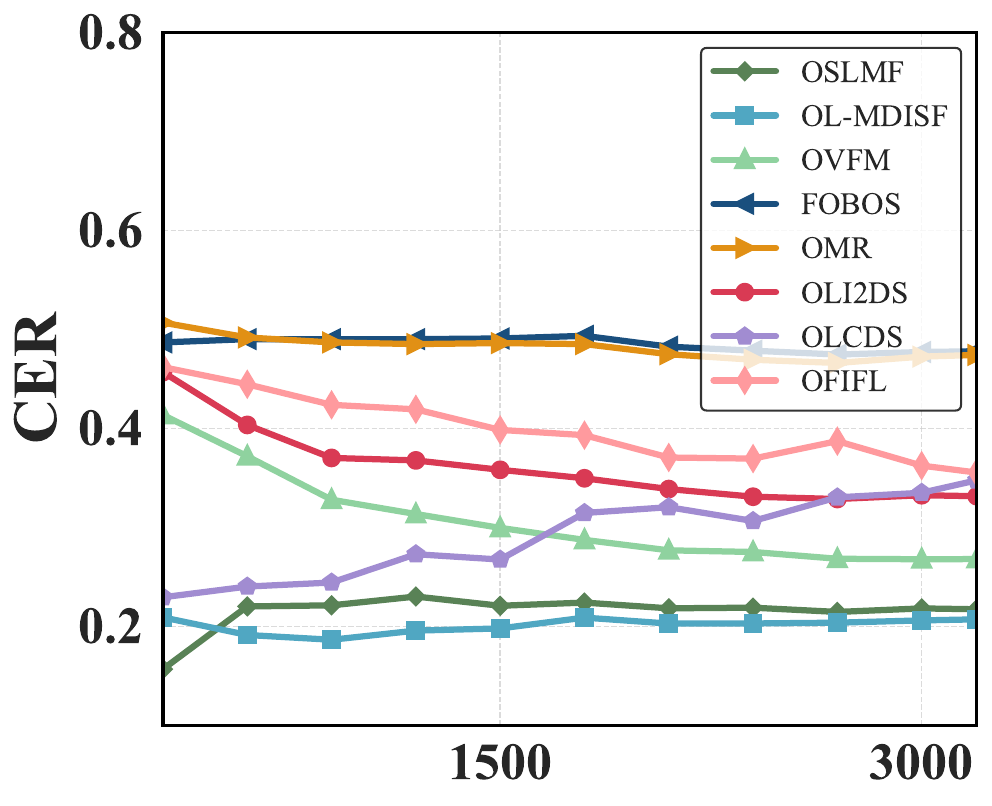}
		\caption{kr-vs-kp}
		\label{fig:Carp_kr-vs-kp_4_CER}
	\end{subfigure}
	\hspace{2em}
	\begin{subfigure}[t]{0.25\linewidth}
		\includegraphics[width=\textwidth]{CER-Capr/Carp_splice.pdf}
		\caption{magic04}
		\label{fig:Carp_magic04_4_CER}
        \vspace{0.5em}
	\end{subfigure}
	\begin{subfigure}[t]{0.25\linewidth}
		\includegraphics[width=\textwidth]{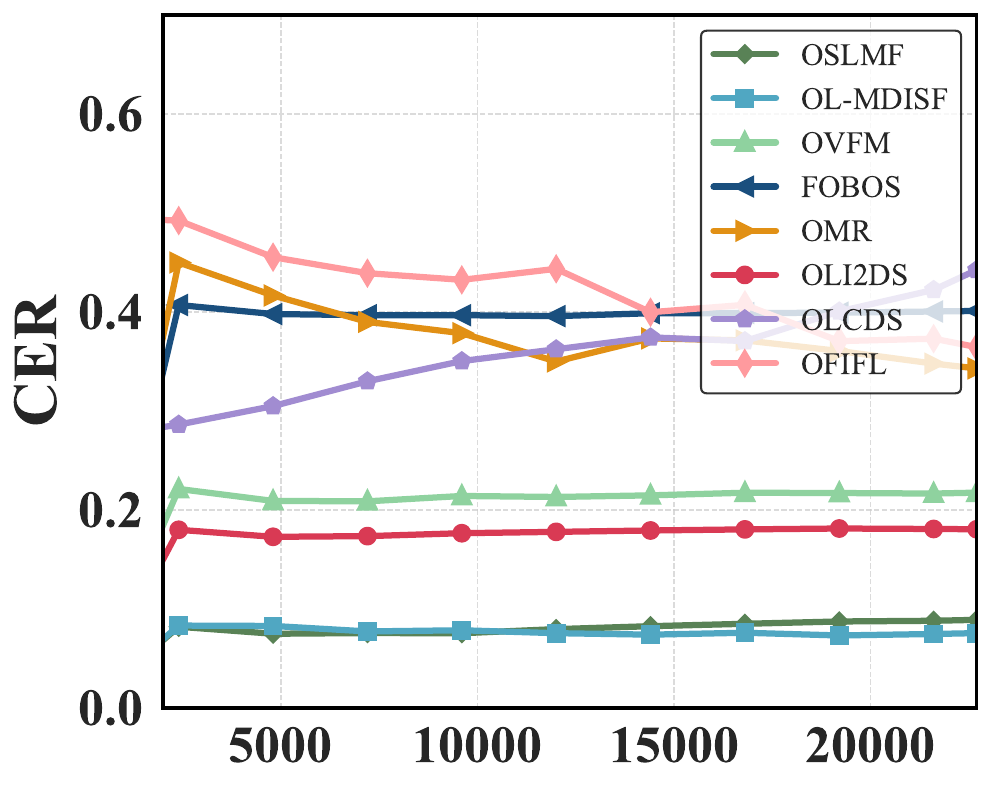}
		\caption{a8a}
		\label{fig:Carp_a8a_4_CER}
	\end{subfigure}
	\hspace{2em}
	\begin{subfigure}[t]{0.25\linewidth}
		\includegraphics[width=\textwidth]{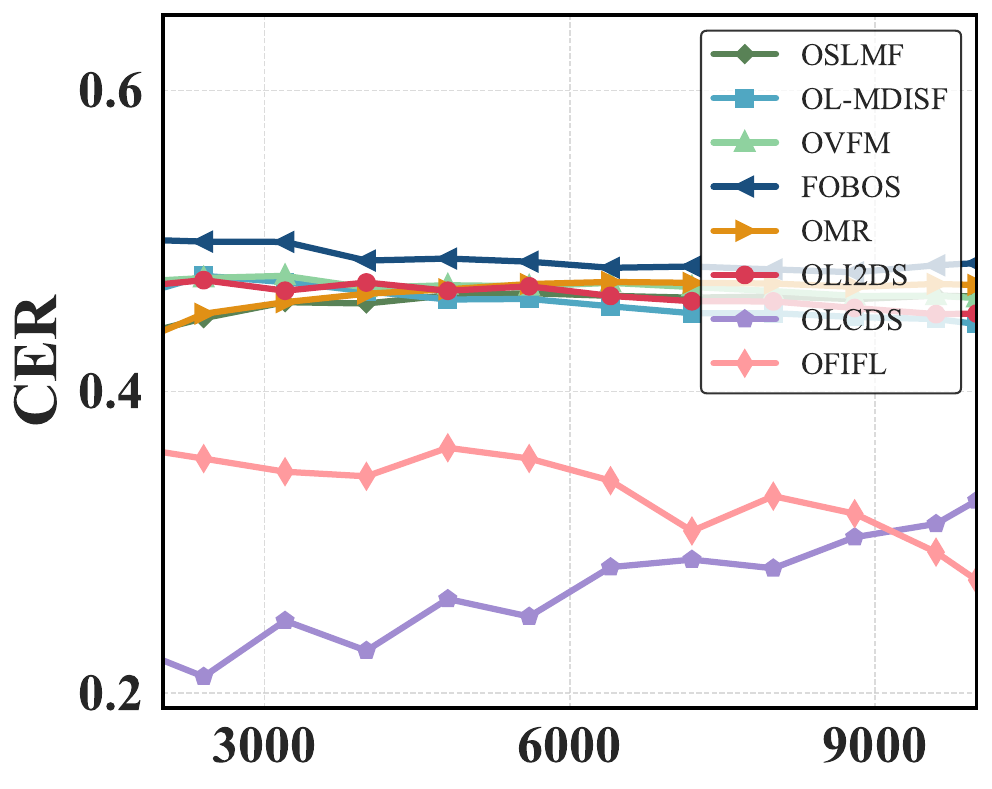}
		\caption{stream}
		\label{fig:Carp_stream_4_CER}
        \vspace{0.5em}
	\end{subfigure}
	\caption{
        The cumulative error rate (CER) trends of OSLMF, \alg,~OVFM, OLI2DS,  
        FOBOS, and OMR in all 14 capricious data streams.
	}
	\label{fig:Capr_4_CER}
\end{figure*}

\begin{figure*}[!t]
	\centering
	\begin{subfigure}[t]{0.25\linewidth}
		\includegraphics[width=\textwidth]{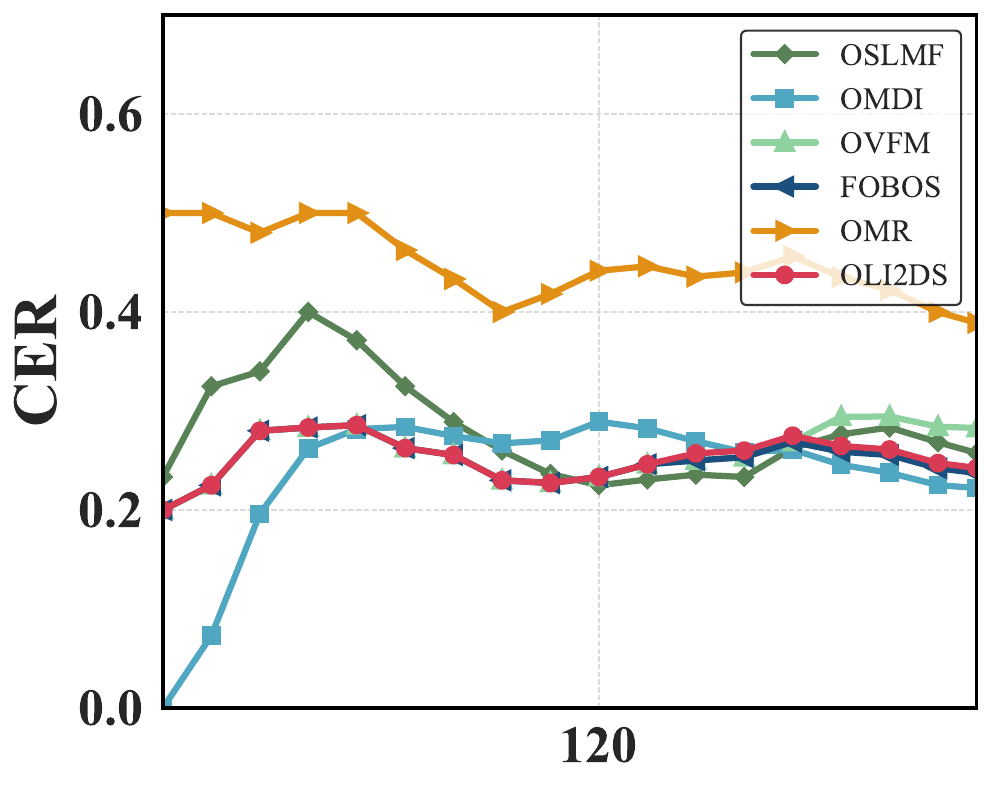}
		\caption{wpbc}
		\label{fig:Trap_wpbc_4_CER}
	\end{subfigure}
	\hspace{2em}
	\begin{subfigure}[t]{0.25\linewidth}
		\includegraphics[width=\textwidth]{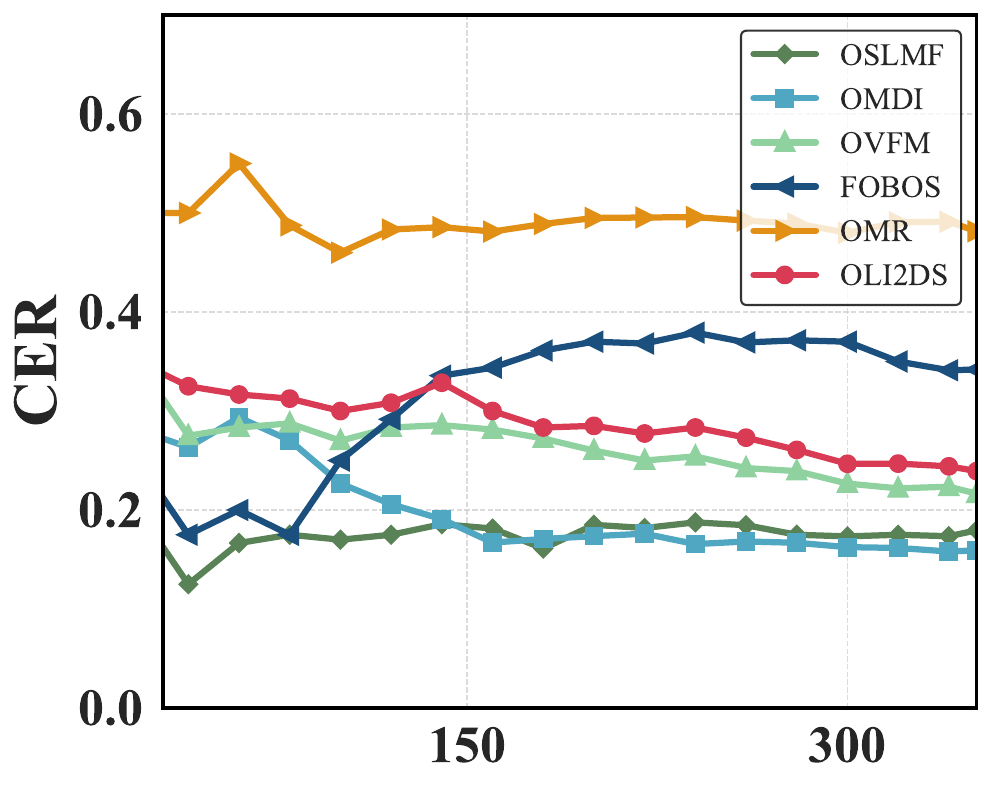}
		\caption{ionosphere}
		\label{fig:Trap_ionosphere_4_CER}
	\end{subfigure}
	\hspace{2em}
	\begin{subfigure}[t]{0.25\linewidth}
		\includegraphics[width=\textwidth]{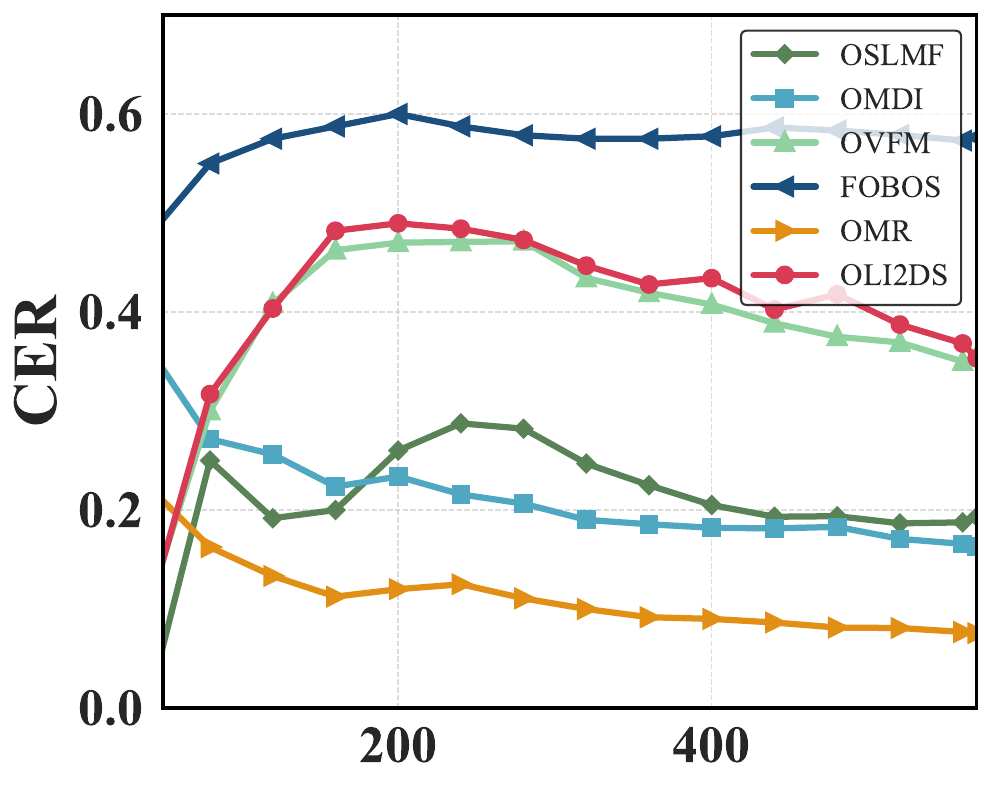}
		\caption{wdbc}
		\label{fig:Trap_wdbc_4_CER}
        \vspace{0.5em}
	\end{subfigure}

	\begin{subfigure}[t]{0.25\linewidth}
		\includegraphics[width=\textwidth]{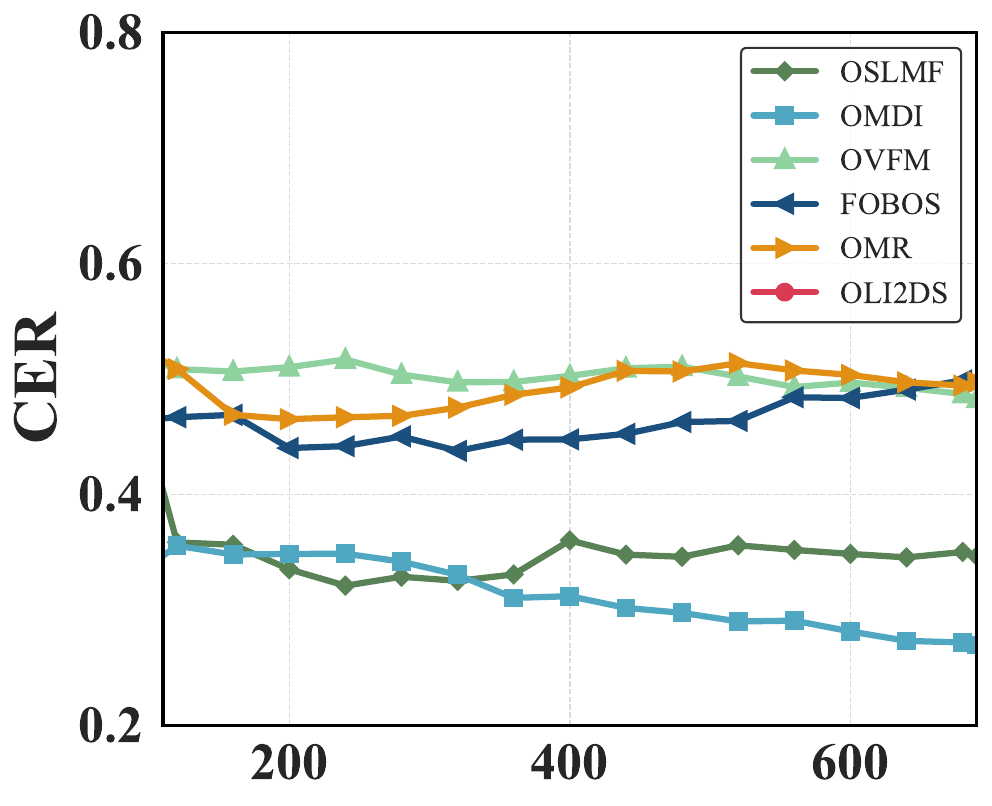}
		\caption{australian}
		\label{fig:Trap_australian_4_CER}
	\end{subfigure}
	\hspace{2em}
	\begin{subfigure}[t]{0.25\linewidth}
		\includegraphics[width=\textwidth]{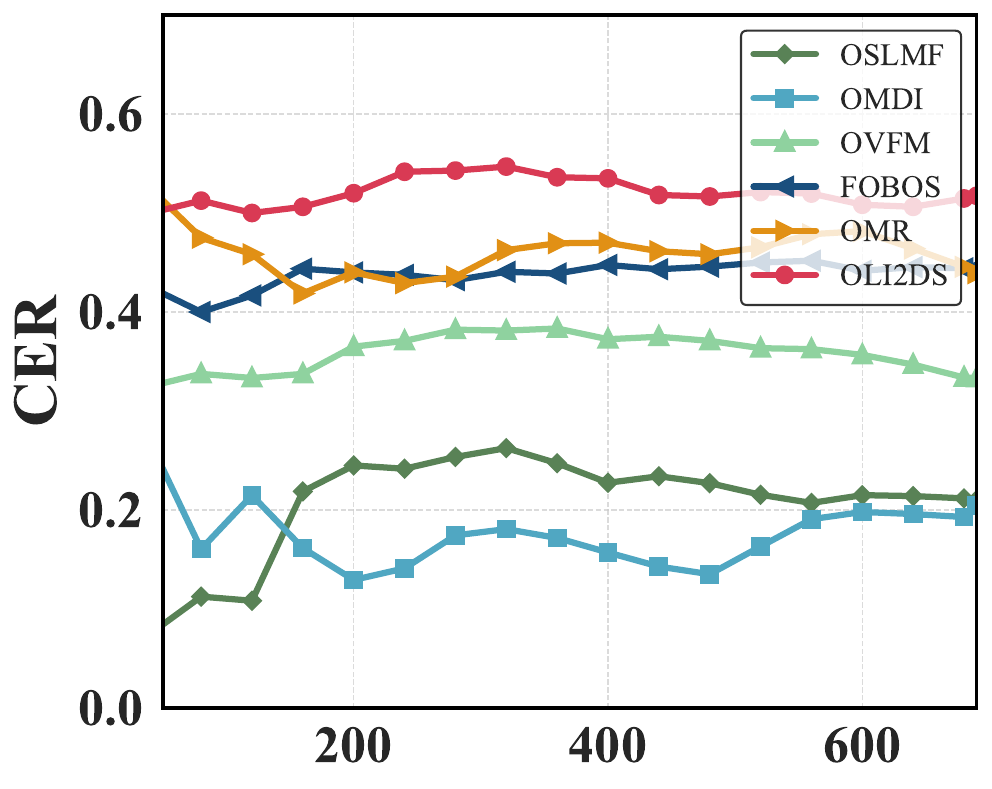}
		\caption{credit-a}
		\label{fig:Trap_credit_4_CER}
	\end{subfigure}
	\hspace{2em}
	\begin{subfigure}[t]{0.25\linewidth}
		\includegraphics[width=\textwidth]{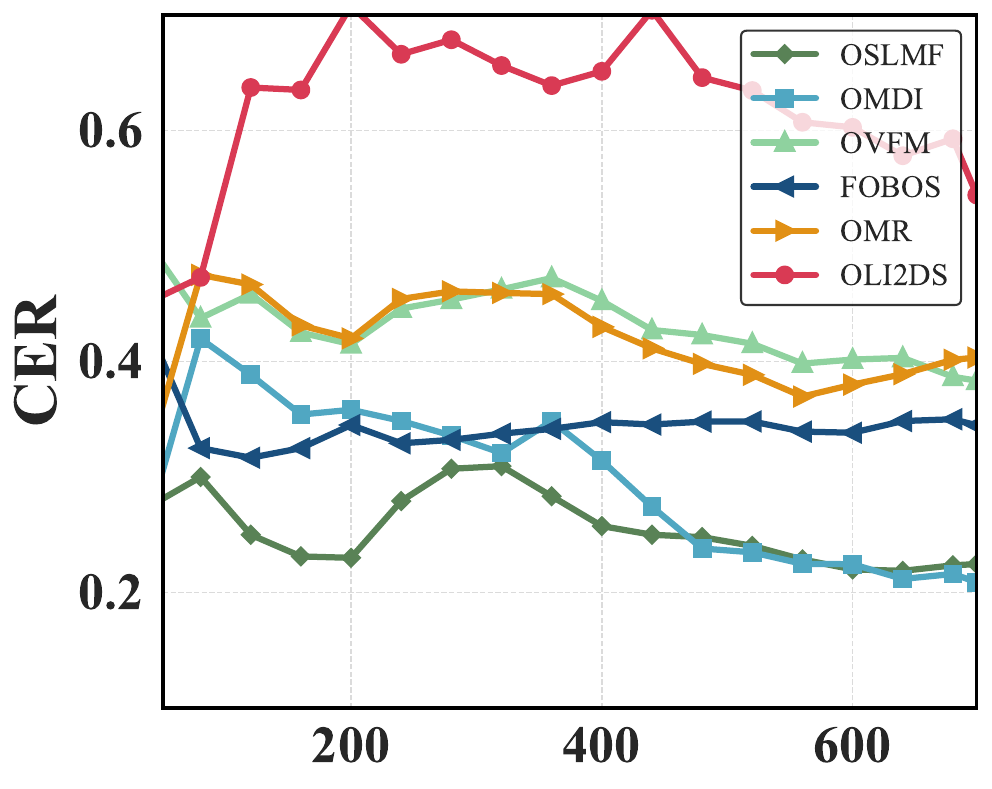}
		\caption{wbc}
		\label{fig:Trap_wbc_4_CER}
        \vspace{0.5em}
	\end{subfigure}

	\begin{subfigure}[t]{0.25\linewidth}
		\includegraphics[width=\textwidth]{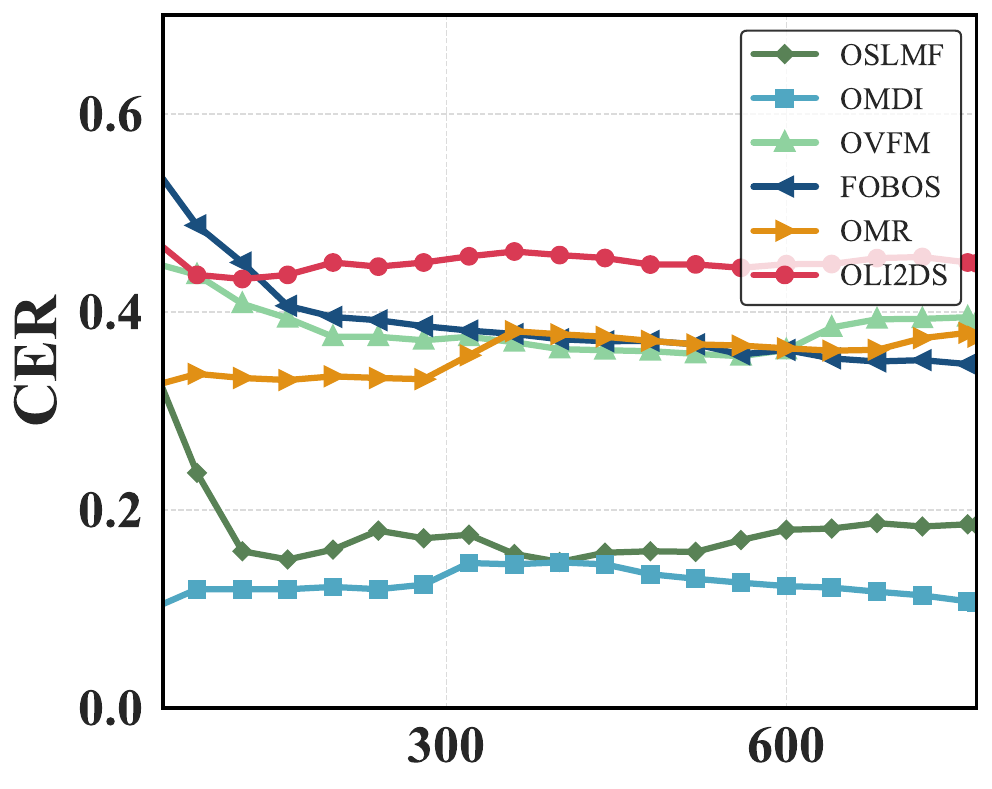}
		\caption{diabetes}
		\label{fig:Trap_diabetes_4_CER}
	\end{subfigure}
	\hspace{2em}
	\begin{subfigure}[t]{0.25\linewidth}
		\includegraphics[width=\textwidth]{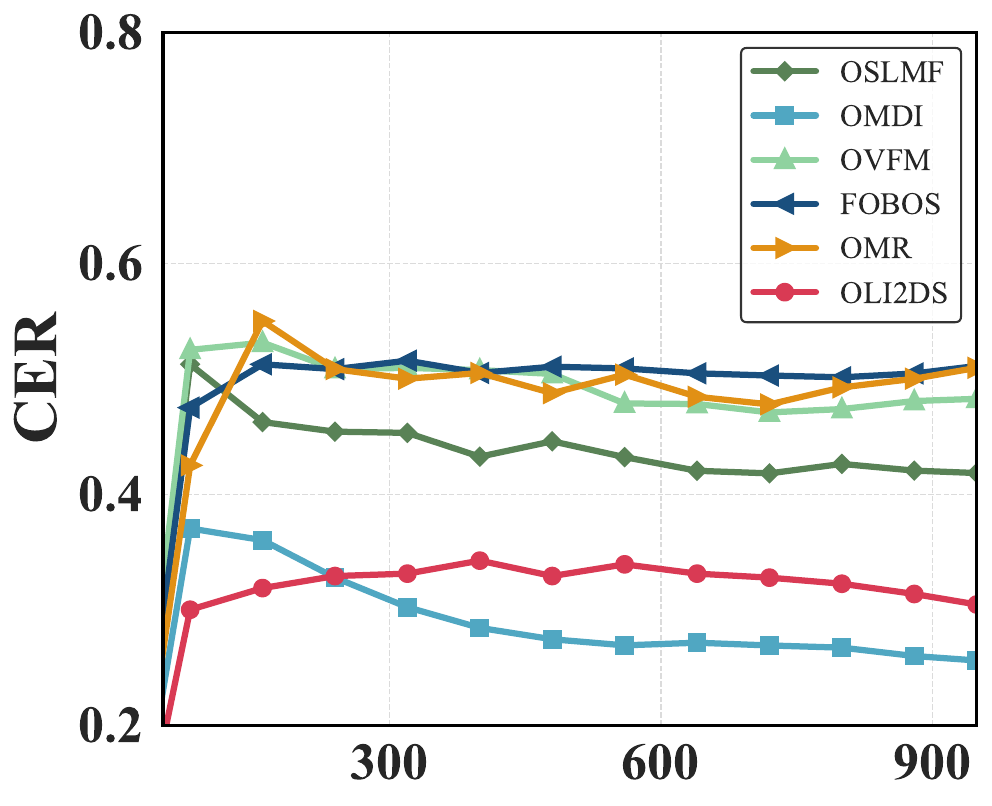}
		\caption{dna}
		\label{fig:Trap_dna_4_CER}
	\end{subfigure}
	\hspace{2em}
	\begin{subfigure}[t]{0.25\linewidth}
		\includegraphics[width=\textwidth]{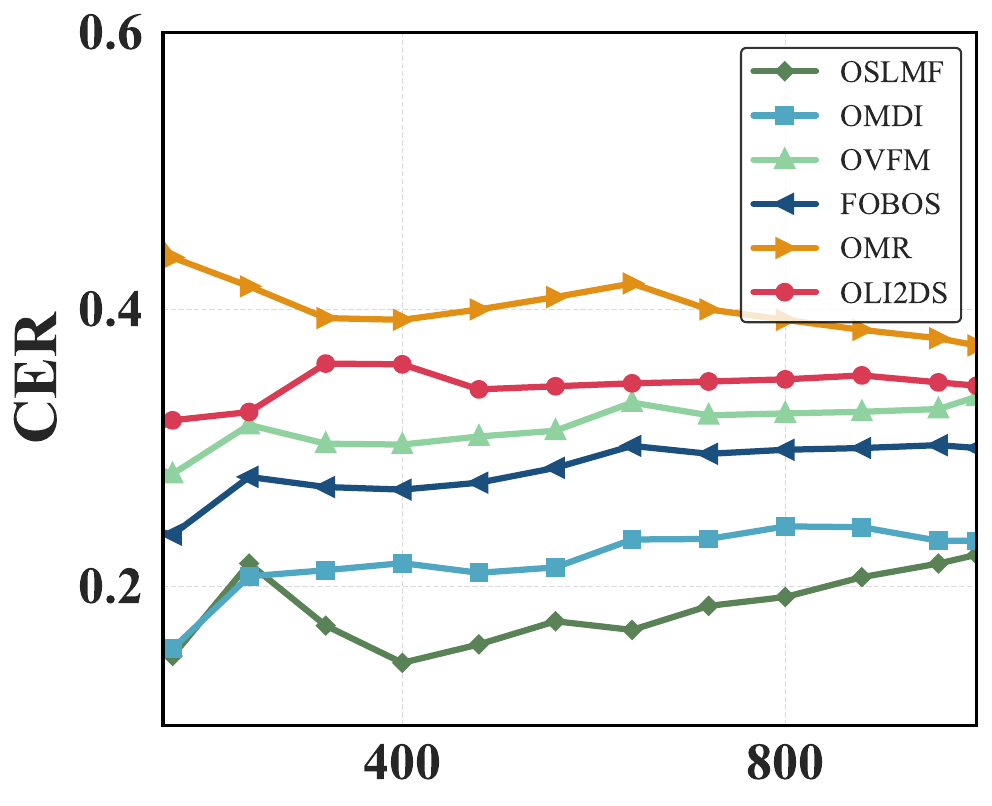}
		\caption{german}
		\label{fig:Trap_german_4_CER}
        \vspace{0.5em}
	\end{subfigure}

	\begin{subfigure}[t]{0.25\linewidth}
		\includegraphics[width=\textwidth]{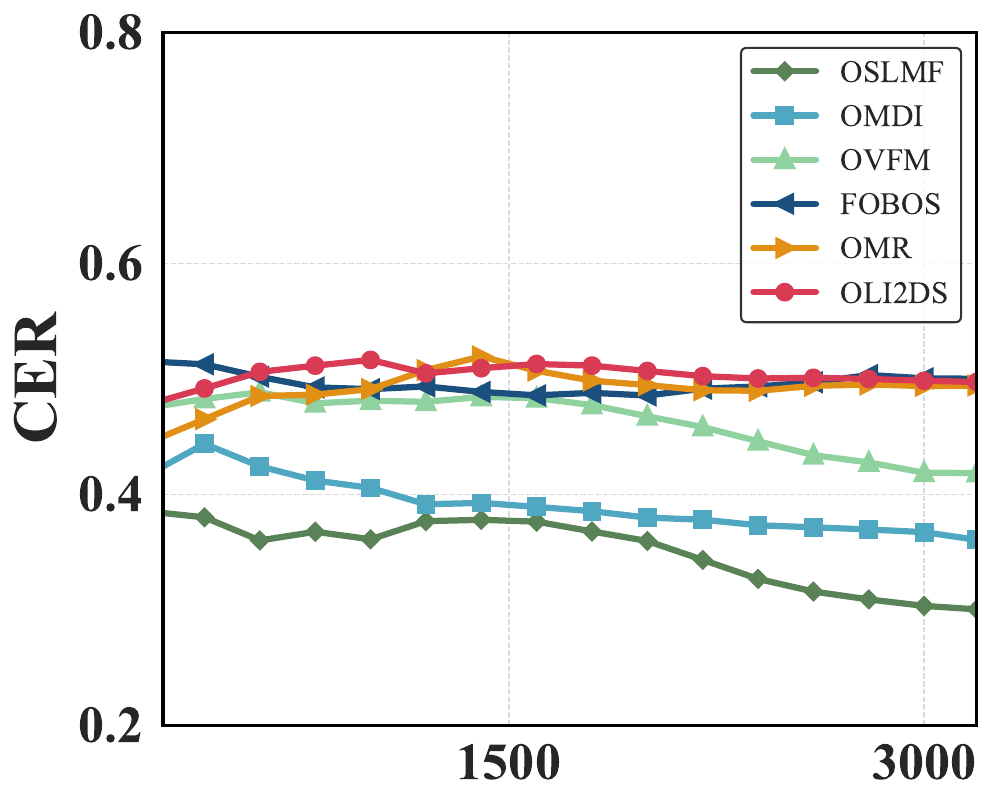}
		\caption{splice}
		\label{fig:Trap_splice_4_CER}
	\end{subfigure}
	\hspace{2em}
	\begin{subfigure}[t]{0.25\linewidth}
		\includegraphics[width=\textwidth]{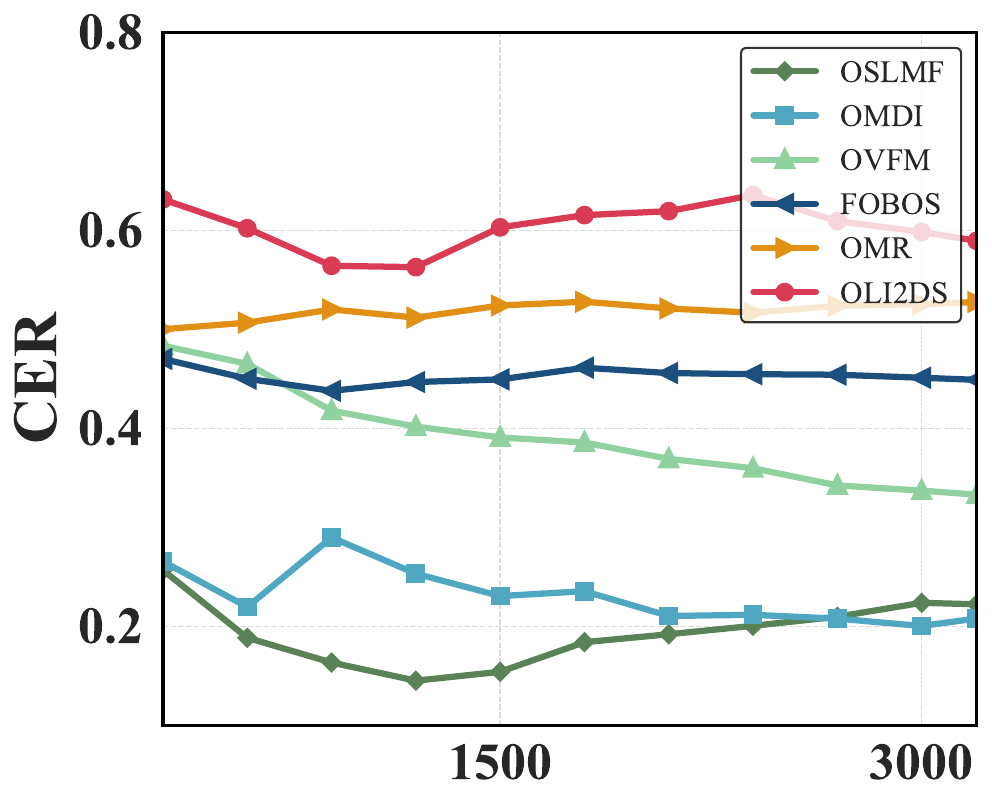}
		\caption{kr-vs-kp}
		\label{fig:Trap_kr-vs-kp_4_CER}
	\end{subfigure}
	\hspace{2em}
	\begin{subfigure}[t]{0.25\linewidth}
		\includegraphics[width=\textwidth]{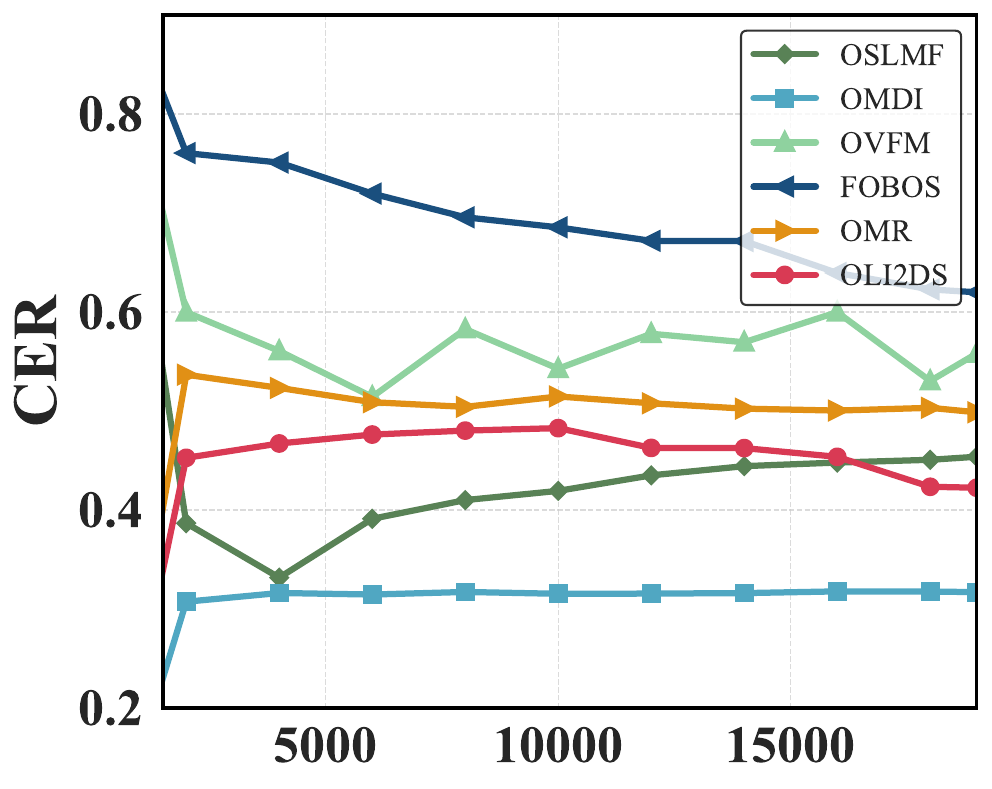}
		\caption{magic04}
		\label{fig:Trap_magic04_4_CER}
        \vspace{0.5em}
	\end{subfigure}

	\begin{subfigure}[t]{0.25\linewidth}
		\includegraphics[width=\textwidth]{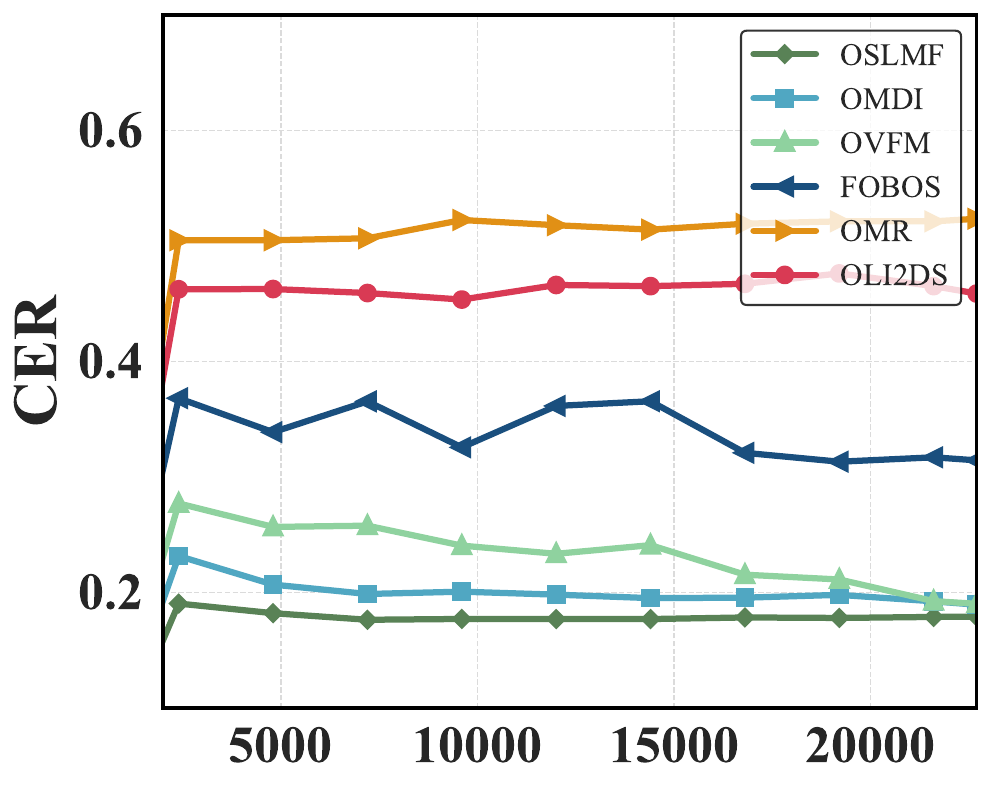}
		\caption{a8a}
		\label{fig:Trap_a8a_4_CER}
	\end{subfigure}
	\hspace{2em}
	\begin{subfigure}[t]{0.25\linewidth}
		\includegraphics[width=\textwidth]{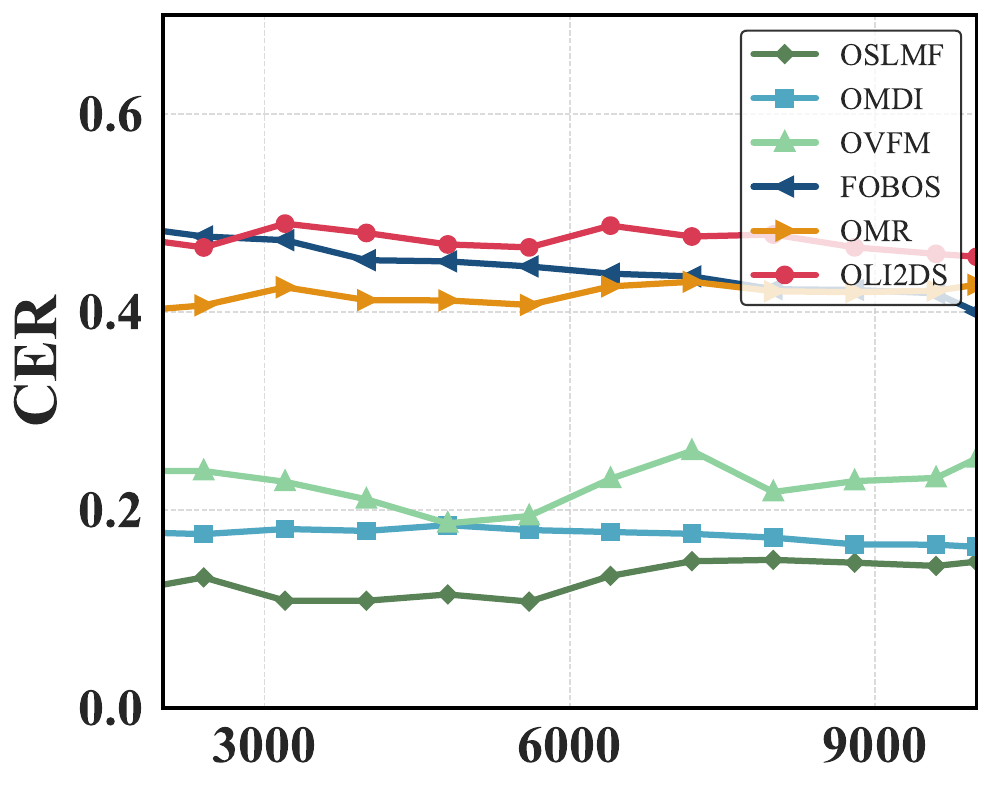}
		\caption{stream}
		\label{fig:Trap_stream_4_CER}
        \vspace{0.5em}
	\end{subfigure}

	\caption{
        The cumulative error rate (CER) trends of OSLMF, \alg,~OLSF, OLI2DS, FOBOS, and OMR in all 14 trapezoidal data streams.
	}
	\label{fig:Trap_4_CER}
\end{figure*}

This section presents additional experimental results that were precluded from the main paper because of page limitations.
These results include:
\begin{enumerate}[leftmargin=*]
	\item The complete CER trends 
	yielded from all 14 datasets, shown in Fig.~\ref{fig:Capr_4_CER},
	which supplements Fig. 4 in the main content. The results are collected from experiments in the setting of capricious data streams.

	\item The CER trends 
	yielded from all 14 datasets, shown in Fig.~\ref{fig:Trap_4_CER},
	which are collected from the setting of trapezoidal data streams.

	\item The complete trends of the temporal variation 
	of ensemble weight $\alpha_1$ and the CERs of 
	\alg\ and its ablation variant \alg-F and \alg-L from all 14 datasets.
	The results supplement Fig. 4 in the main content,
	which are in the setting of capricious data streams.
	\item The complete trends of the temporal variation 
	of ensemble weight $\alpha_1$ and the CERs of 
	\alg\  and its ablation variant \alg-F and \alg-L from all 14 datasets
	in the setting of trapezoidal data streams.

        \item  Multiple different label missing ratios were set for various base classifiers to perform a two-dimensional model performance comparison (Table \ref{tab:1missing}).
\end{enumerate}

\begin{figure*}[!t]
	\centering
	\begin{subfigure}[t]{0.27\linewidth}
		\includegraphics[width=\textwidth]{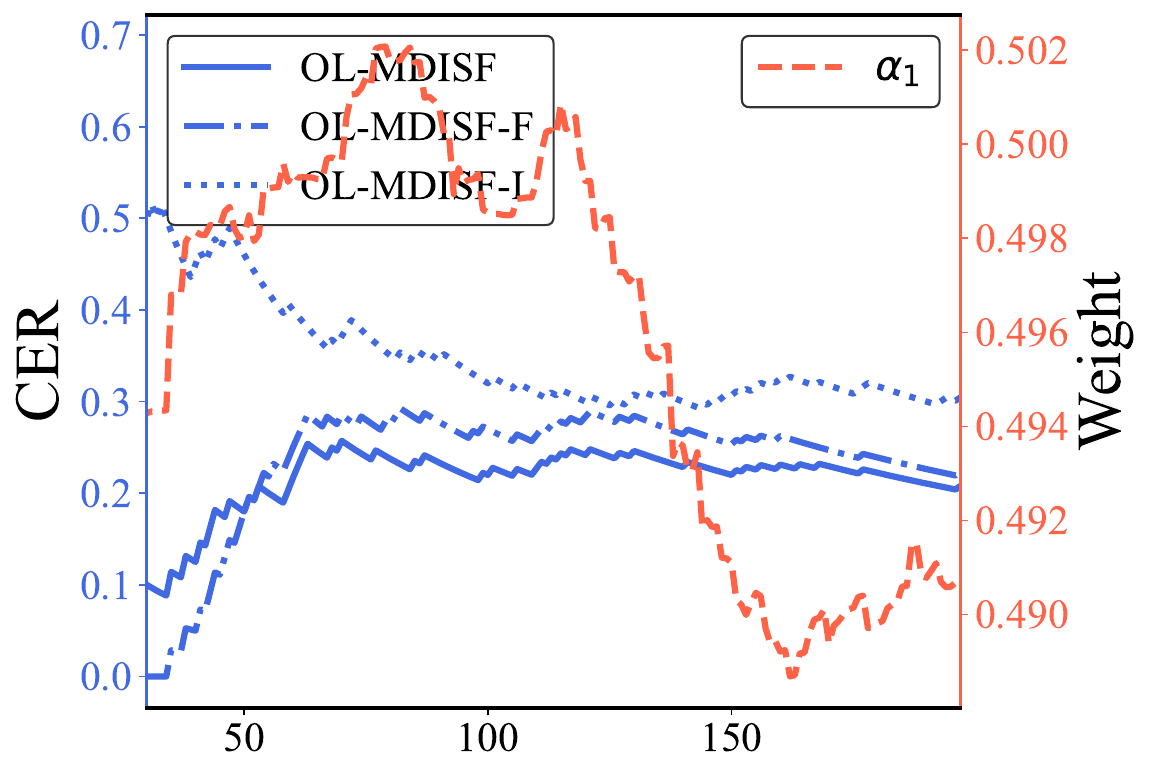}
		\caption{wpbc}
		\label{fig:carp_wpbc_CER}
	\end{subfigure}
	\hspace{2em}
	\begin{subfigure}[t]{0.27\linewidth}
		\includegraphics[width=\textwidth]{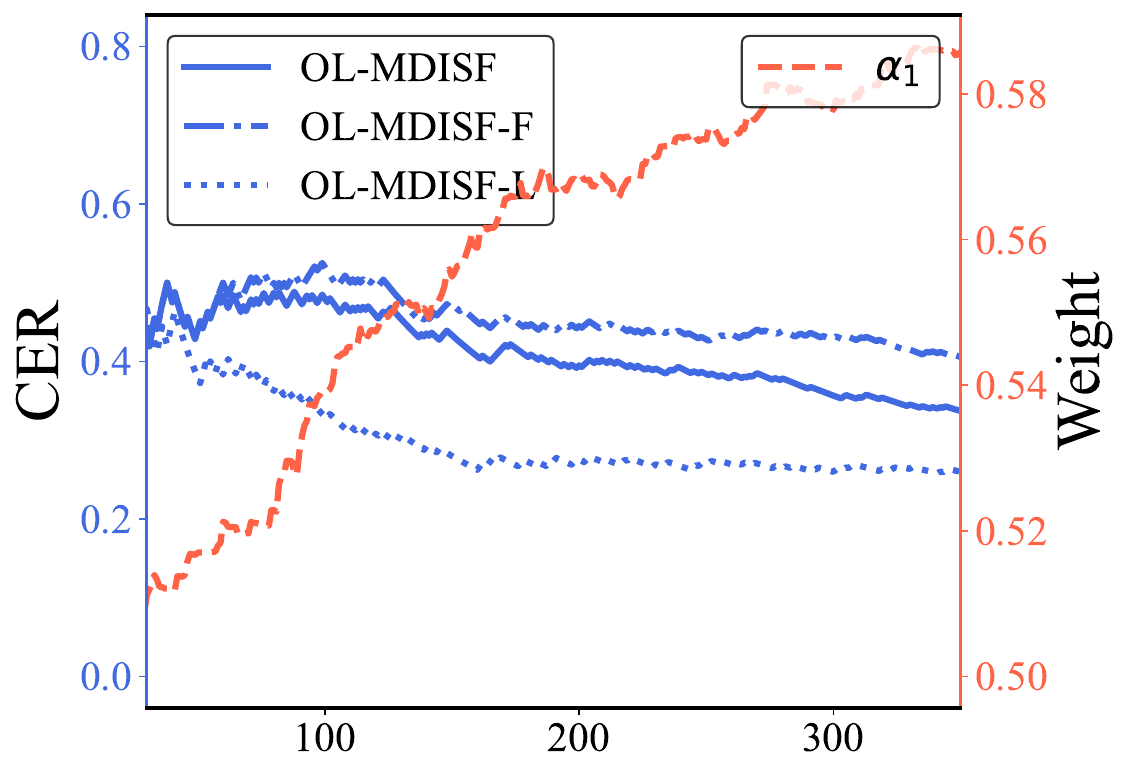}
		\caption{ionosphere}
		\label{fig:carp_ionosphere_CER}
	\end{subfigure}
	\hspace{2em}
	\begin{subfigure}[t]{0.27\linewidth}
		\includegraphics[width=\textwidth]{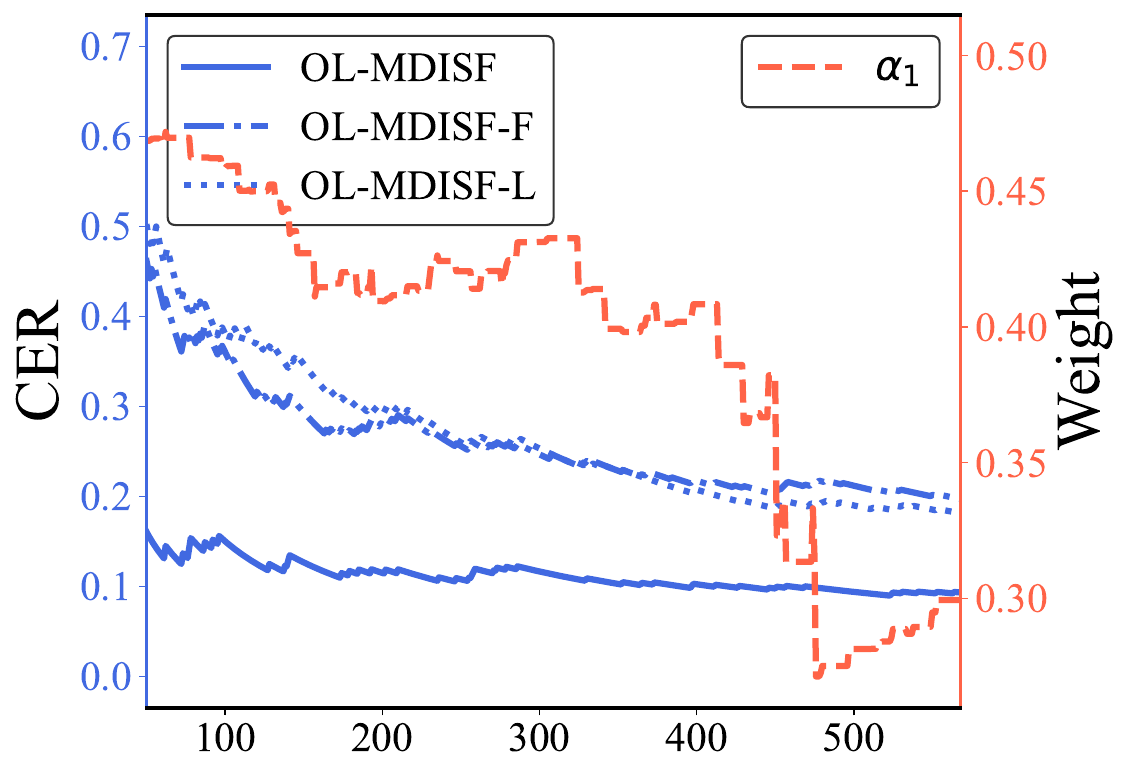}
		\caption{wdbc}
		\label{fig:carp_wdbc_CER}
        \vspace{0.5em}
	\end{subfigure}

	\begin{subfigure}[t]{0.27\linewidth}
		\includegraphics[width=\textwidth]{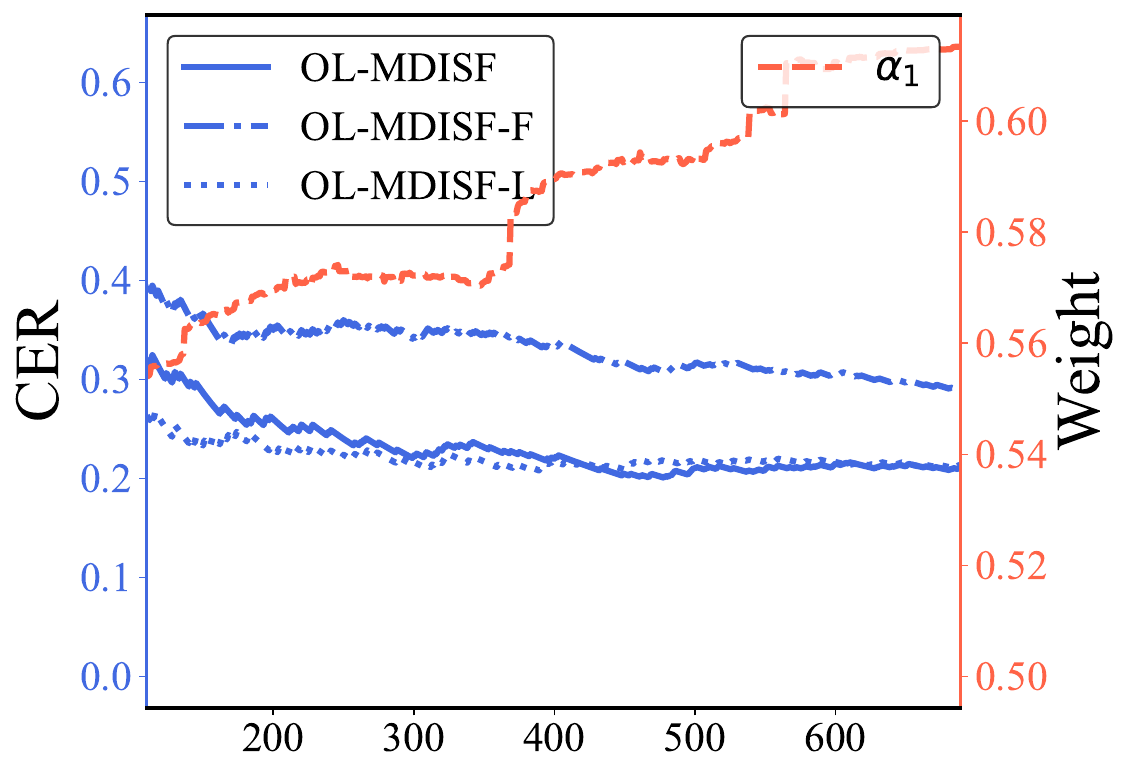}
		\caption{australian}
		\label{fig:carp_australian_CER}
	\end{subfigure}
	\hspace{2em}
	\begin{subfigure}[t]{0.27\linewidth}
		\includegraphics[width=\textwidth]{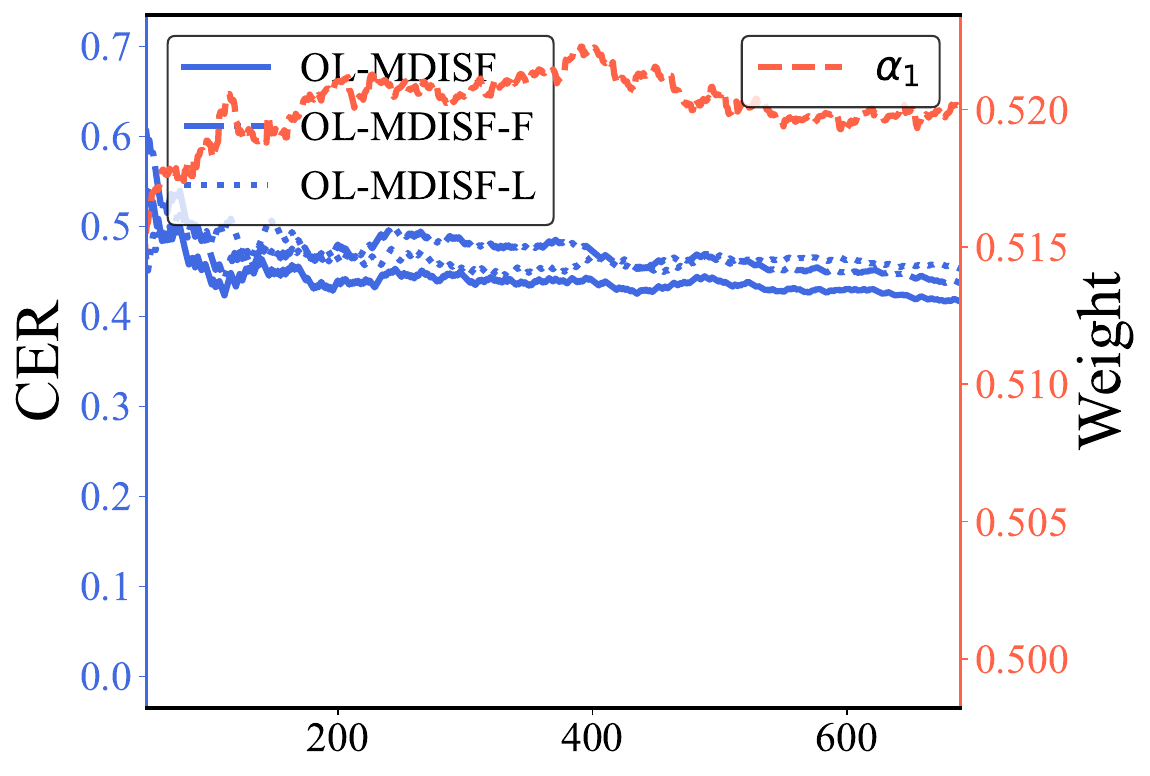}
		\caption{credit-a}
		\label{fig:carp_credit_CER}
	\end{subfigure}
	\hspace{2em}
	\begin{subfigure}[t]{0.27\linewidth}
		\includegraphics[width=\textwidth]{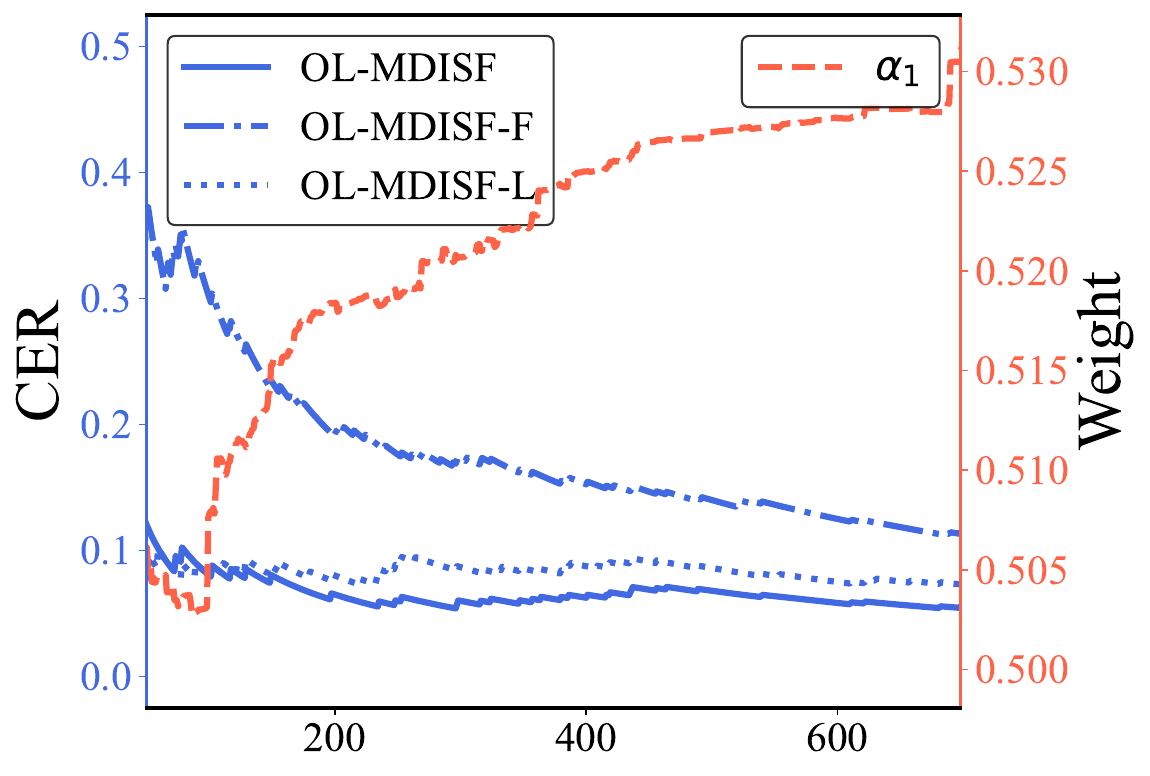}
		\caption{wbc}
		\label{fig:carp_wbc_CER}
        \vspace{0.5em}
	\end{subfigure}

	\begin{subfigure}[t]{0.27\linewidth}
		\includegraphics[width=\textwidth]{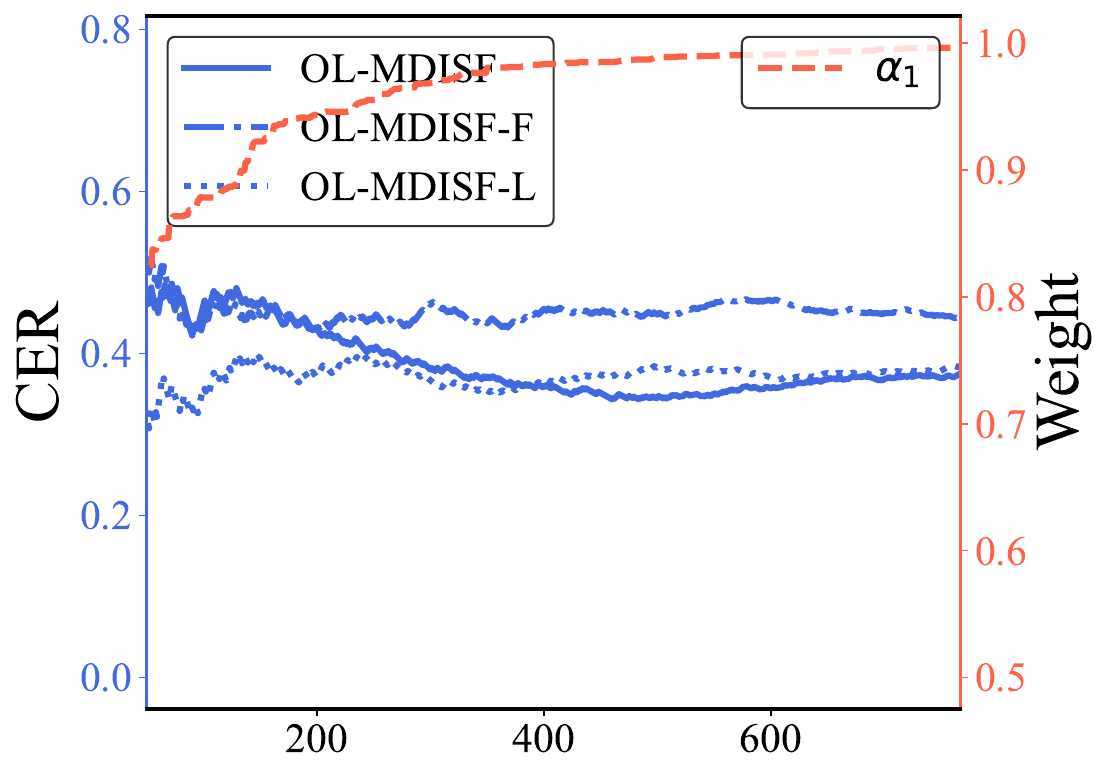}
		\caption{diabetes}
		\label{fig:carp_diabetes_CER}
	\end{subfigure}
	\hspace{2em}
	\begin{subfigure}[t]{0.27\linewidth}
		\includegraphics[width=\textwidth]{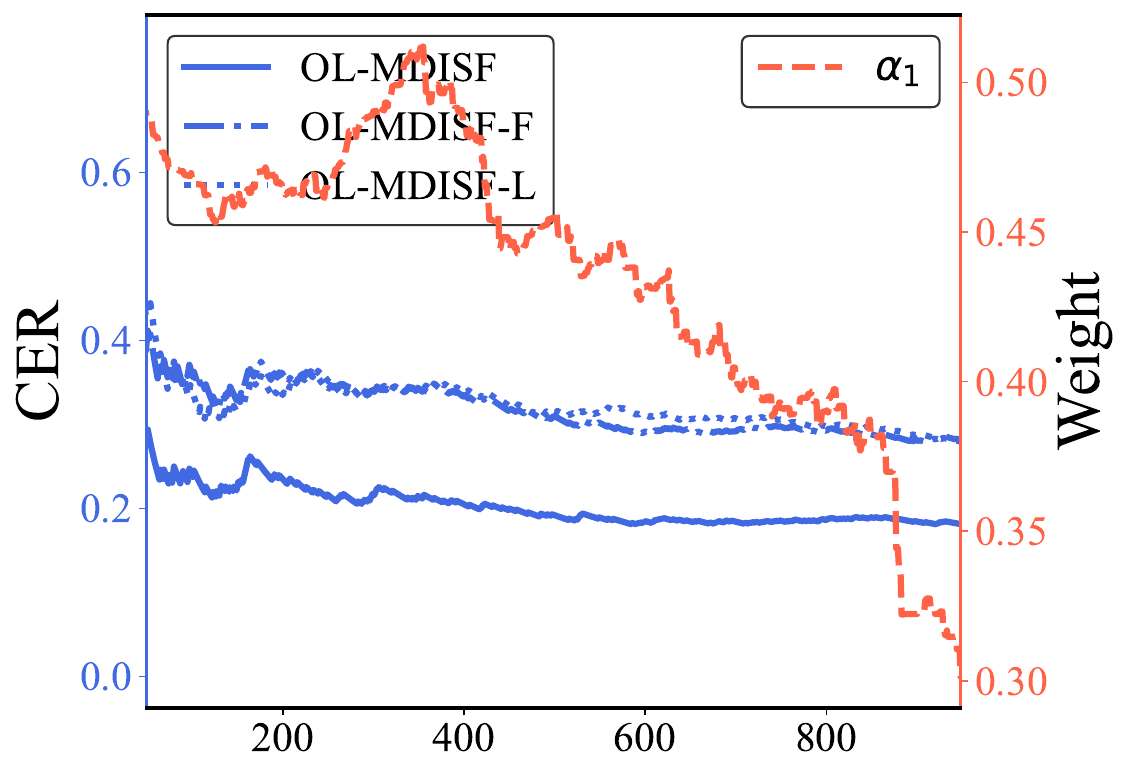}
		\caption{dna}
		\label{fig:carp_dna_CER}
	\end{subfigure}
	\hspace{2em}
	\begin{subfigure}[t]{0.27\linewidth}
		\includegraphics[width=\textwidth]{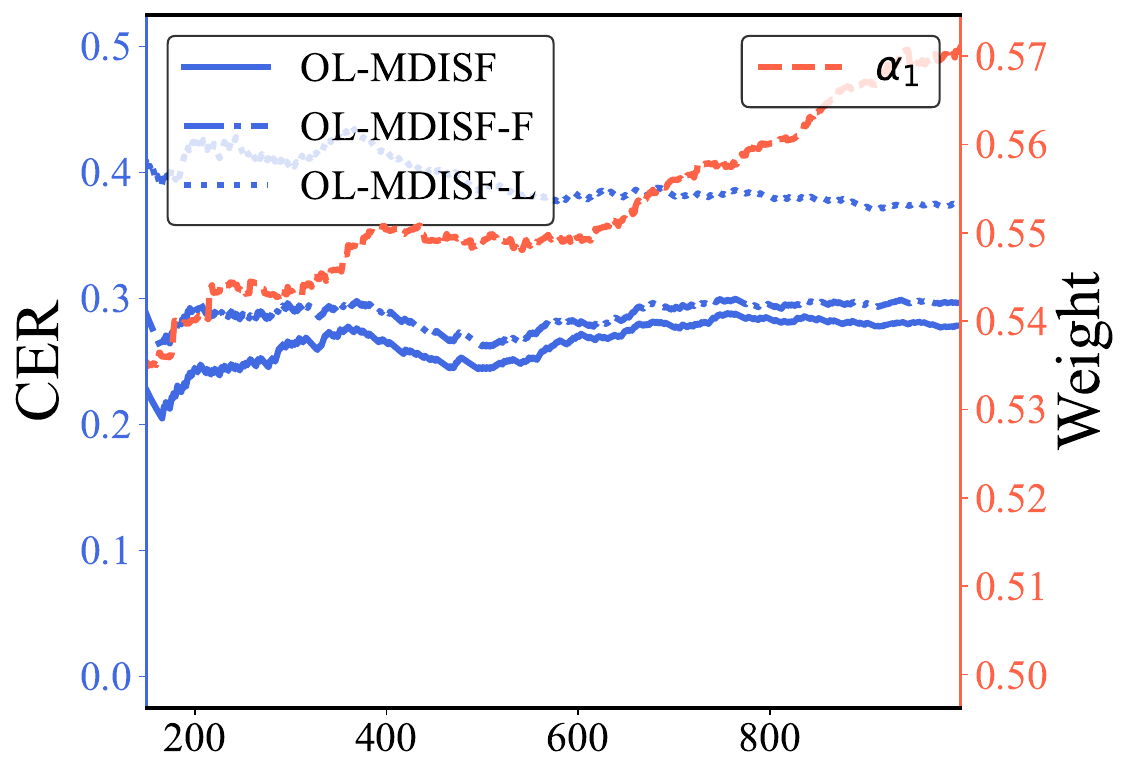}
		\caption{german}
		\label{fig:carp_german_CER}
        \vspace{0.5em}
	\end{subfigure}

	\begin{subfigure}[t]{0.27\linewidth}
		\includegraphics[width=\textwidth]{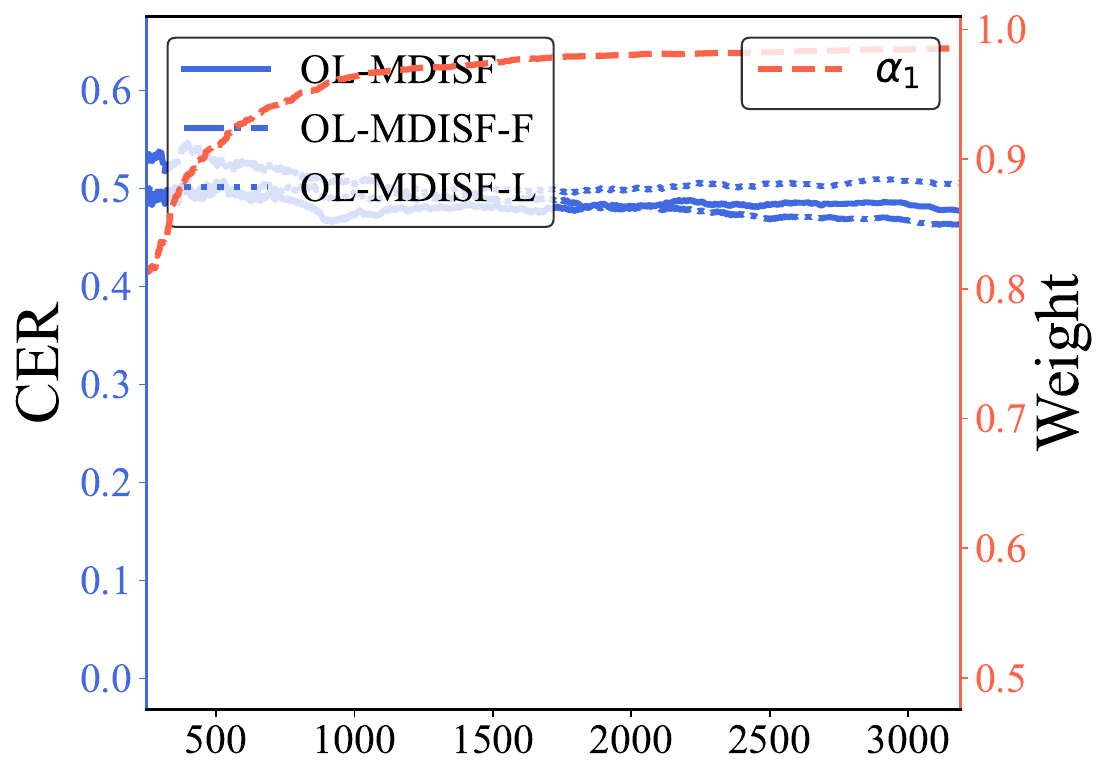}
		\caption{splice}
		\label{fig:carp_splice_CER}
	\end{subfigure}
	\hspace{2em}
	\begin{subfigure}[t]{0.27\linewidth}
		\includegraphics[width=\textwidth]{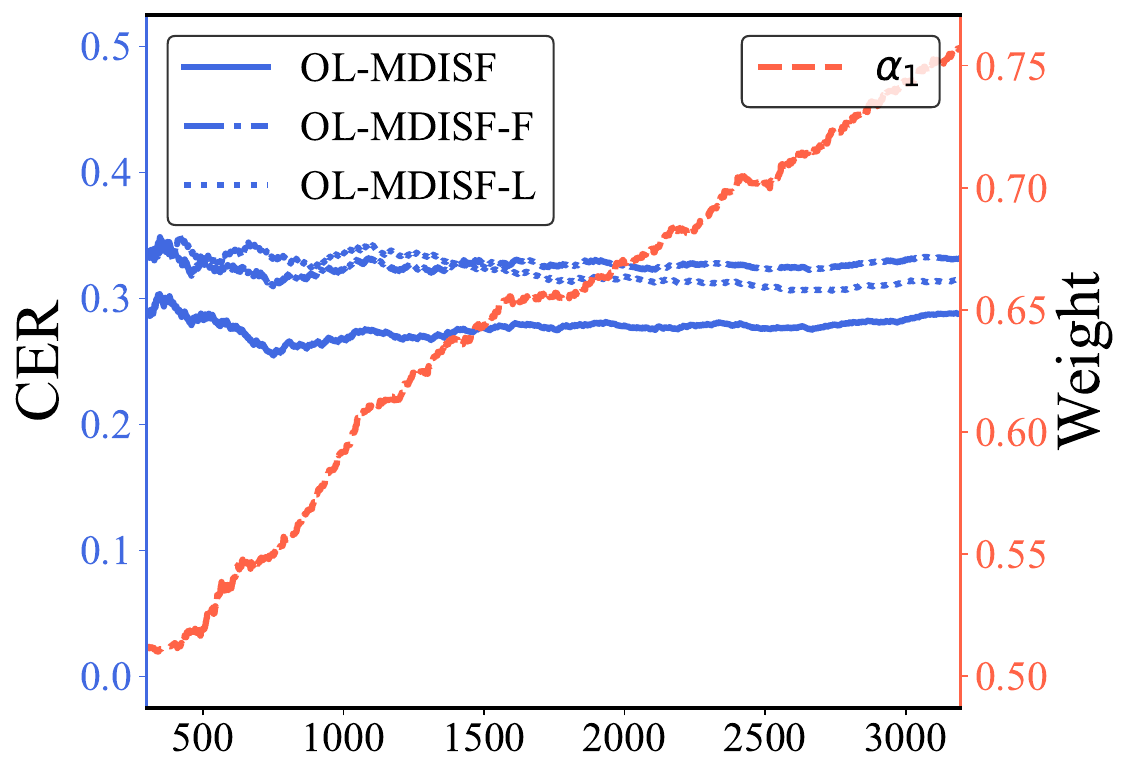}
		\caption{kr-vs-kp}
		\label{fig:carp_kr-vs-kp_CER}
	\end{subfigure}
	\hspace{2em}
	\begin{subfigure}[t]{0.27\linewidth}
		\includegraphics[width=\textwidth]{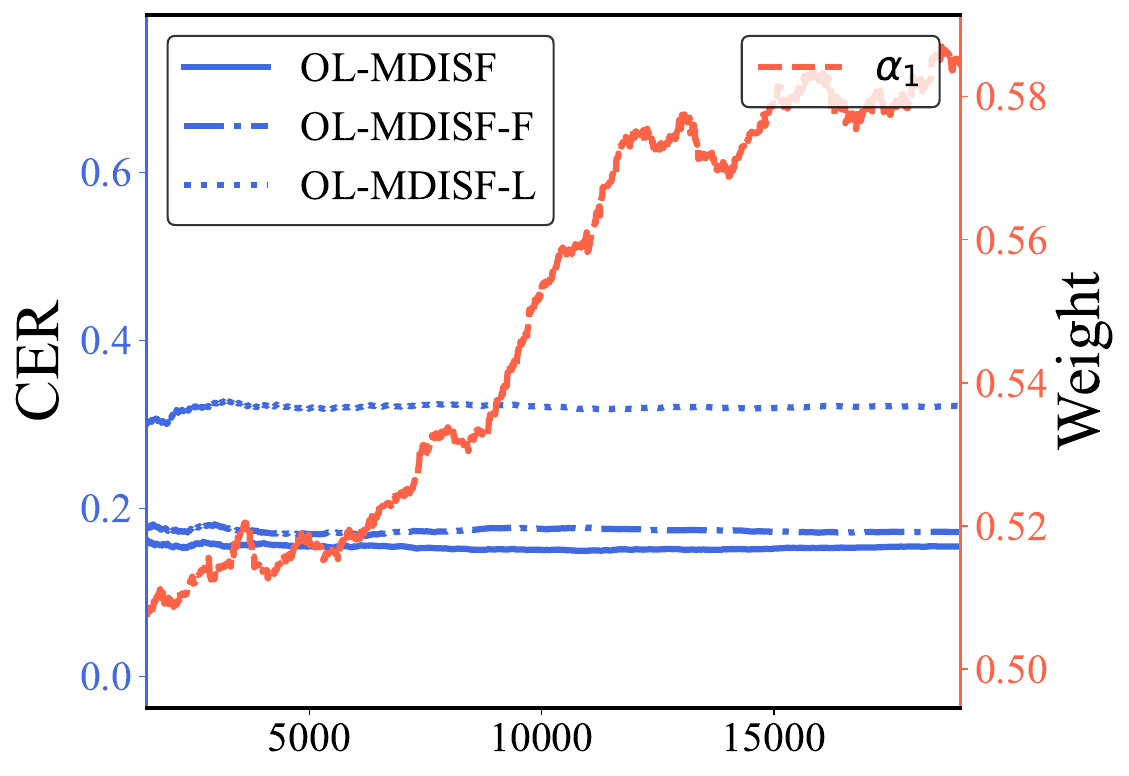}
		\caption{magic04}
		\label{fig:carp_magic04_CER}
        \vspace{0.5em}
	\end{subfigure}

	\begin{subfigure}[t]{0.27\linewidth}
		\includegraphics[width=\textwidth]{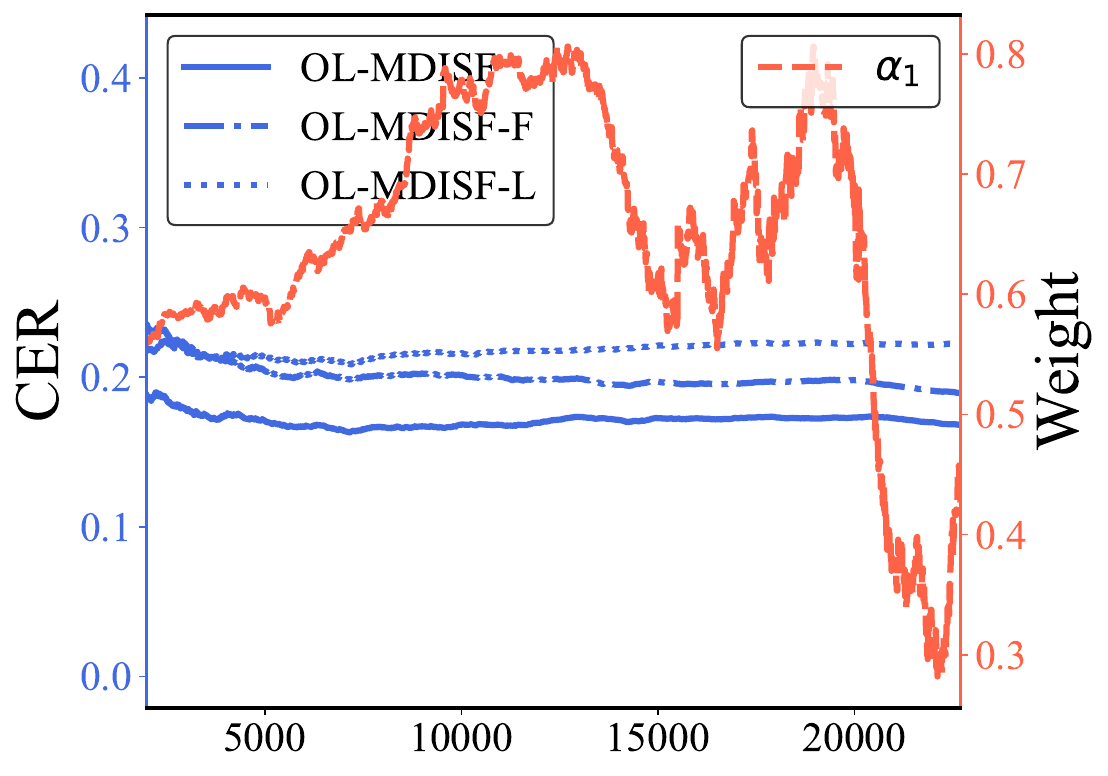}
		\caption{a8a}
		\label{fig:carp_a8a_CER}
	\end{subfigure}
	\hspace{2em}
	\begin{subfigure}[t]{0.27\linewidth}
		\includegraphics[width=\textwidth]{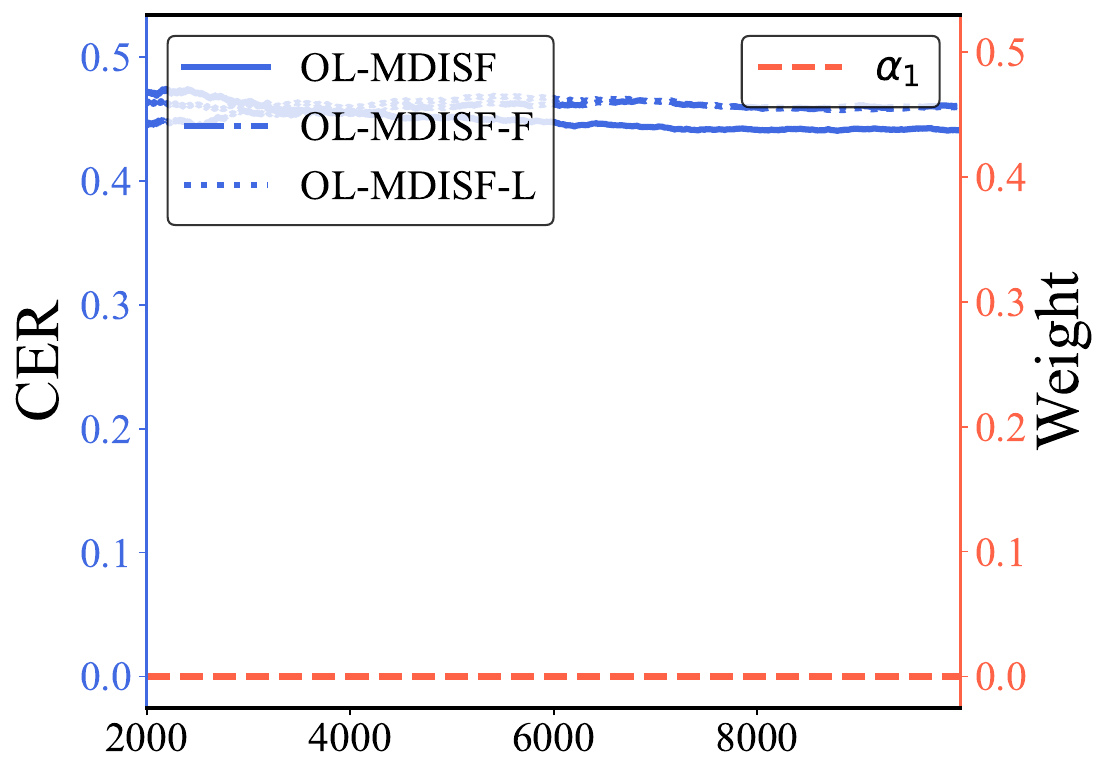}
		\caption{Stream}
		\label{fig:carp_Stream_CER}
        \vspace{0.5em}
	\end{subfigure}

	\caption{
        Temporal variation of ensemble weight $\alpha_1$ and CERs of \alg~and its ablation variant \alg-F and \alg-L in all 14 capricious data streams.
	}
	\label{fig:Cap_alpha_CER}
\end{figure*}

\begin{figure*}[!t]
	\centering
	\begin{subfigure}[t]{0.27\linewidth}
		\includegraphics[width=\textwidth]{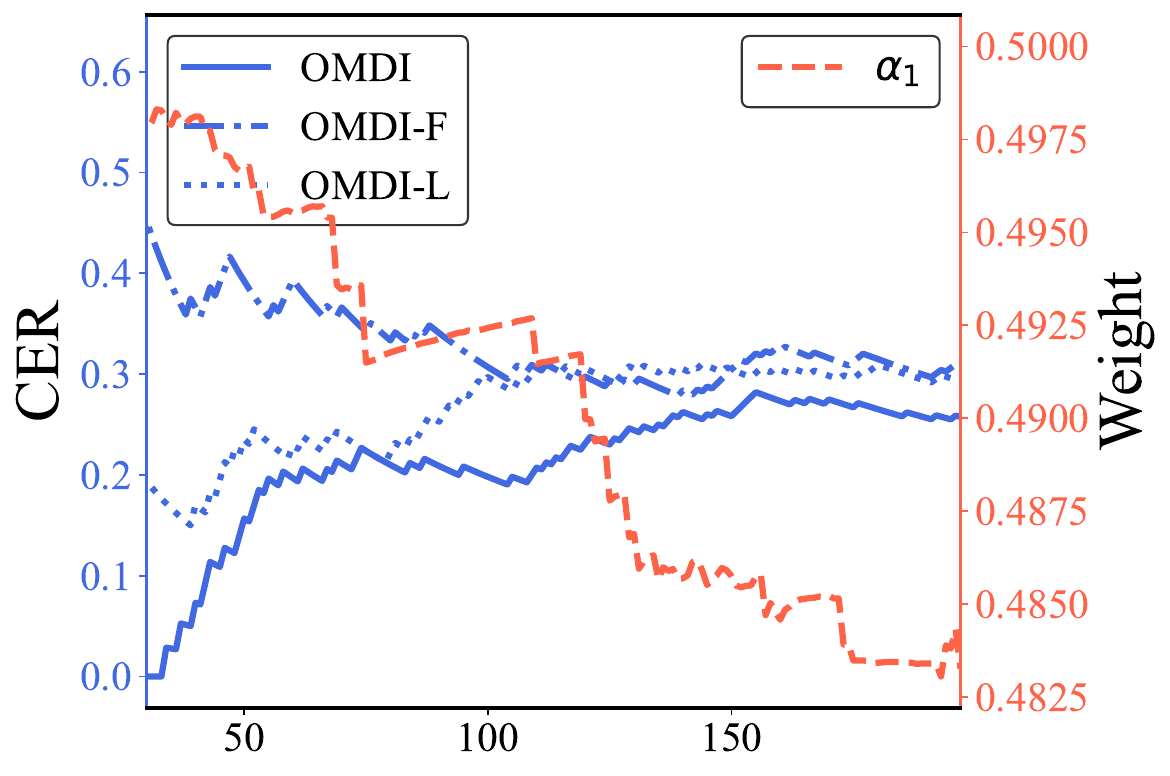}
		\caption{wpbc}
		\label{fig:Trap_wpbc_CER}
	\end{subfigure}
	\hspace{2em}
	\begin{subfigure}[t]{0.27\linewidth}
		\includegraphics[width=\textwidth]{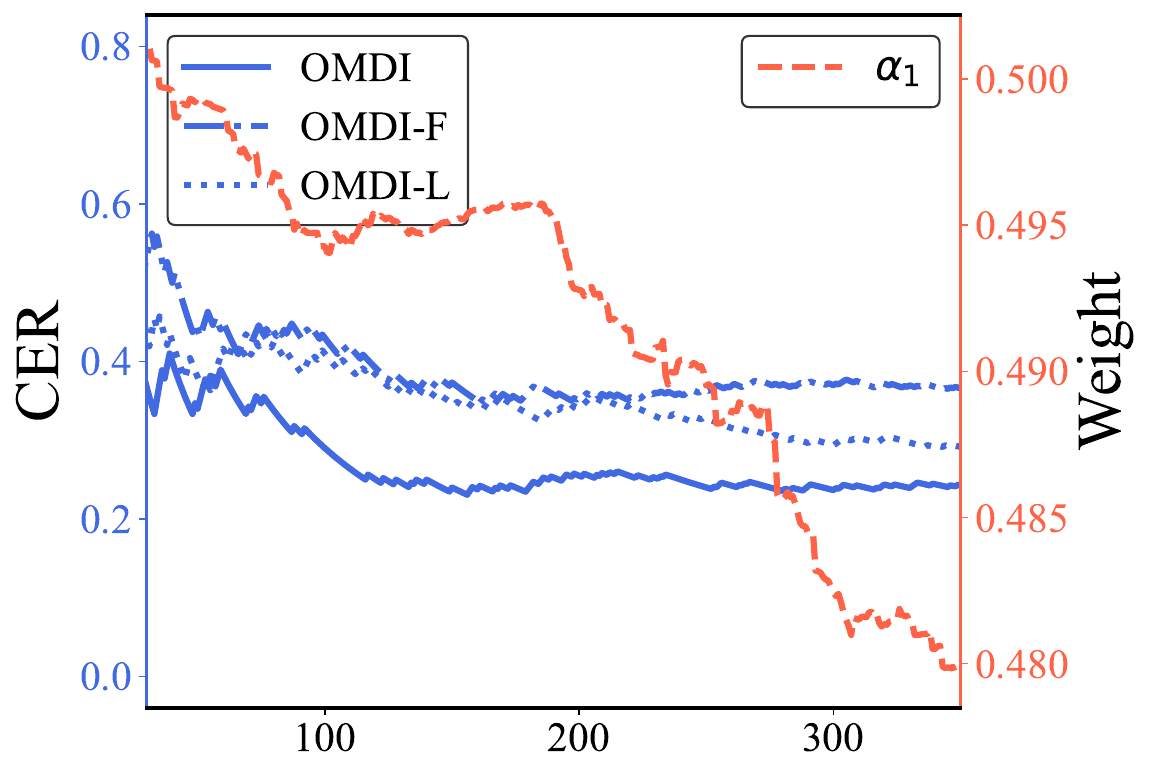}
		\caption{ionosphere}
		\label{fig:Trap_ionosphere_CER}
	\end{subfigure}
	\hspace{2em}
	\begin{subfigure}[t]{0.27\linewidth}
		\includegraphics[width=\textwidth]{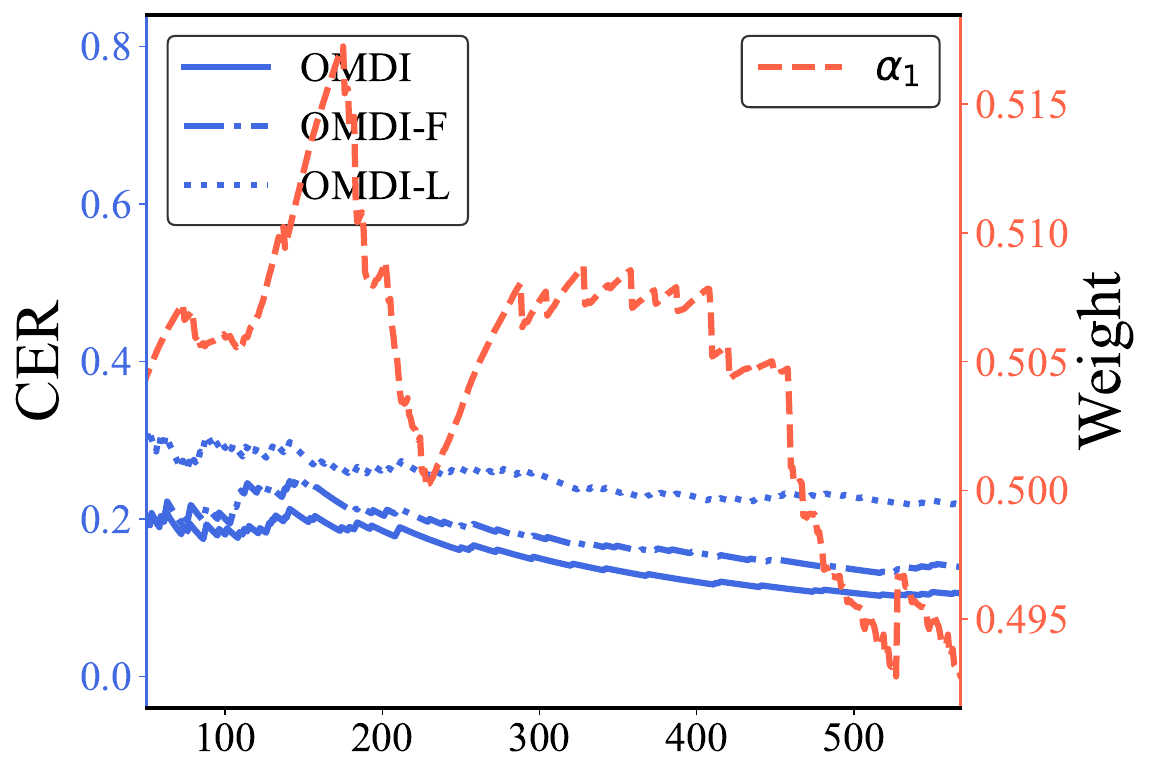}
		\caption{wdbc}
		\label{fig:Trap_wdbc_CER}
        \vspace{0.5em}
	\end{subfigure}

	\begin{subfigure}[t]{0.27\linewidth}
		\includegraphics[width=\textwidth]{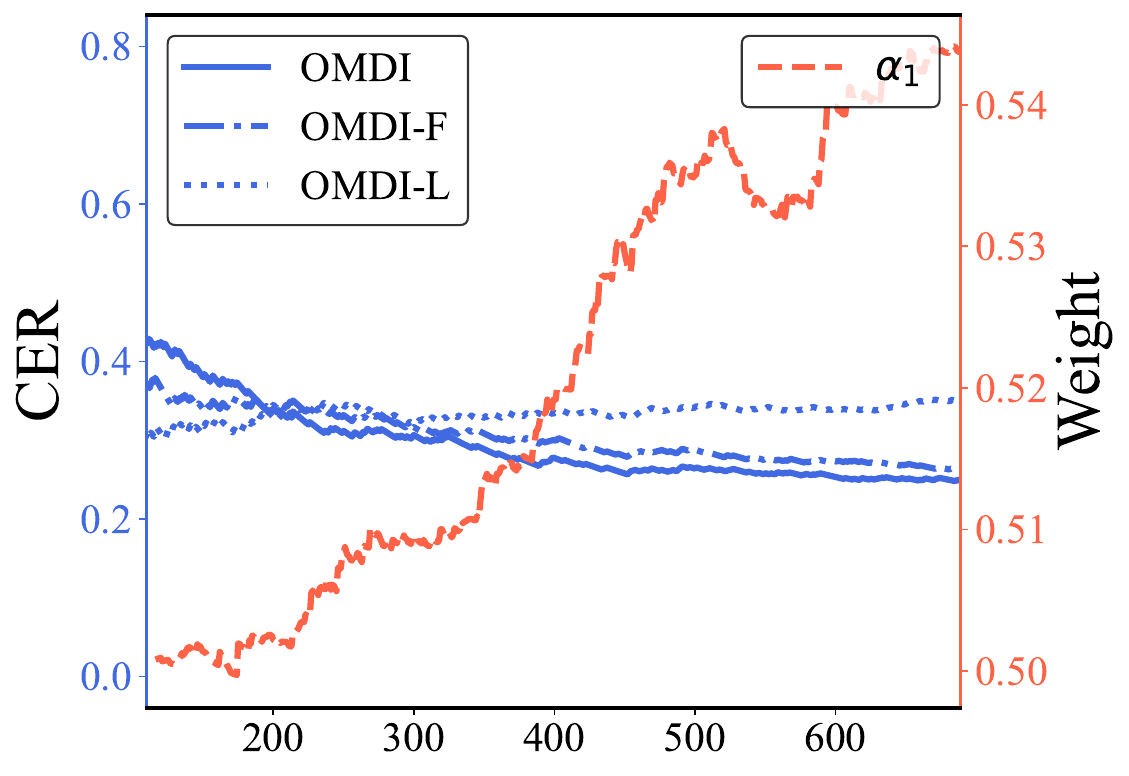}
		\caption{australian}
		\label{fig:Trap_australian_CER}
	\end{subfigure}
	\hspace{2em}
	\begin{subfigure}[t]{0.27\linewidth}
		\includegraphics[width=\textwidth]{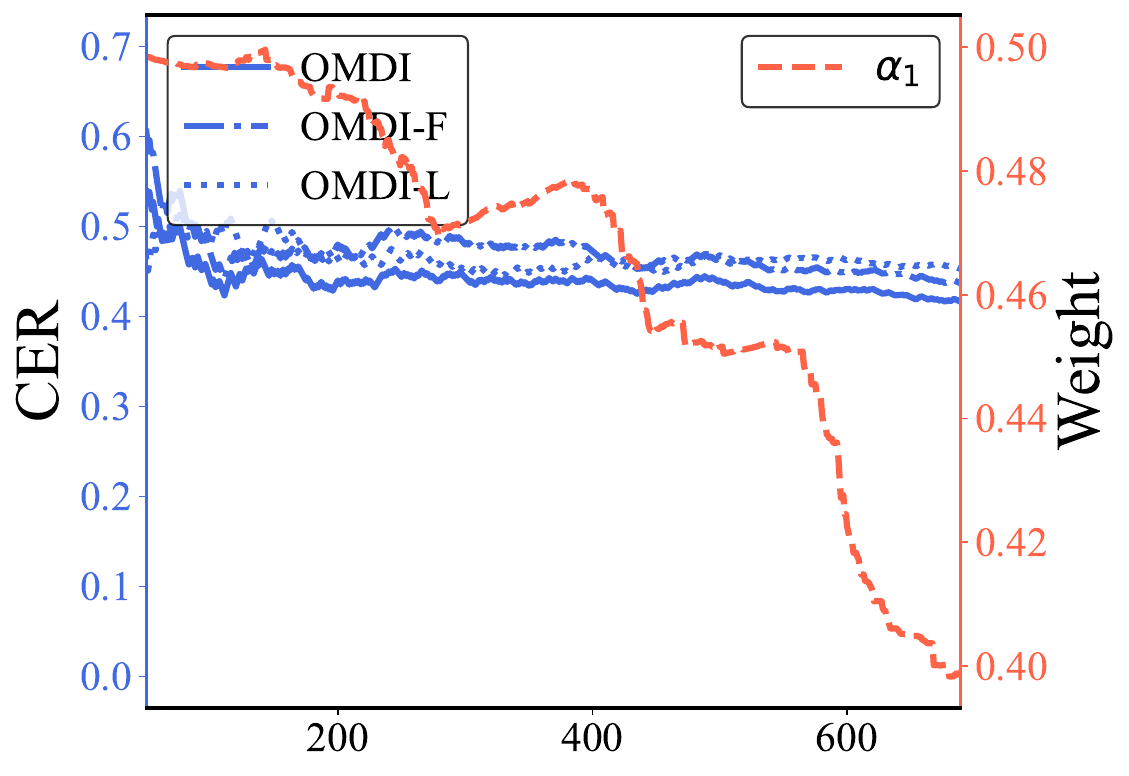}
		\caption{credit-a}
		\label{fig:Trap_credit_CER}
	\end{subfigure}
	\hspace{2em}
	\begin{subfigure}[t]{0.27\linewidth}
		\includegraphics[width=\textwidth]{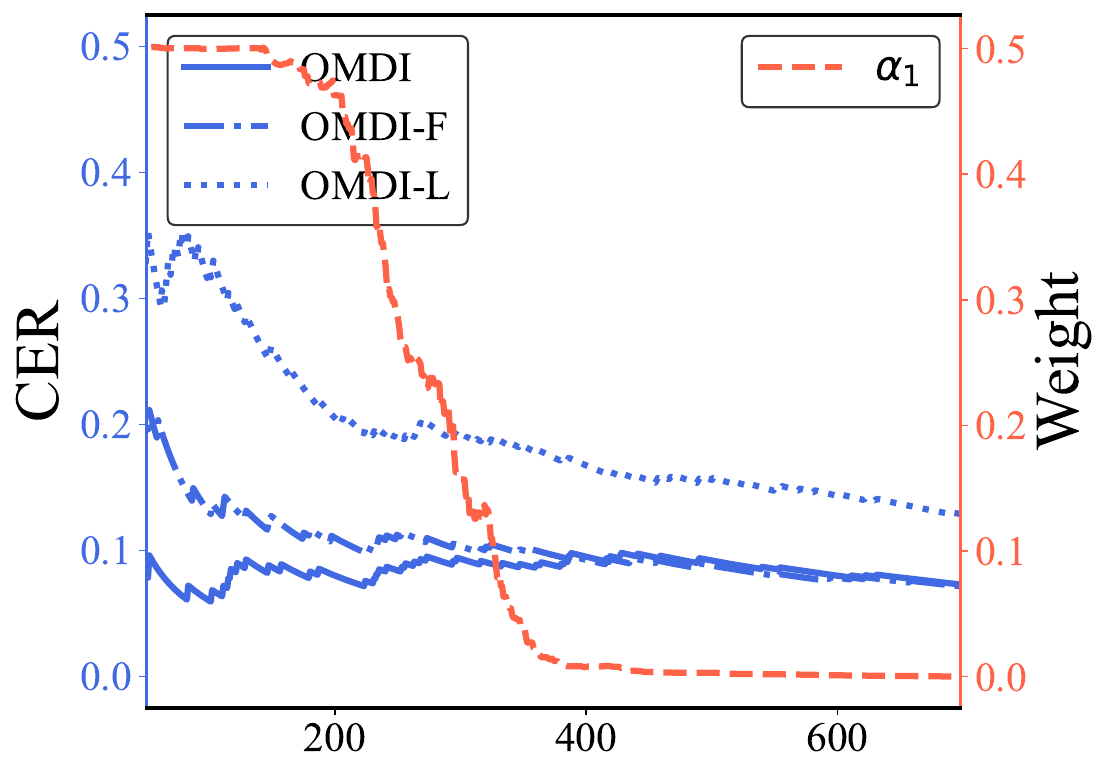}
		\caption{wbc}
		\label{fig:Trap_wbc_CER}
        \vspace{0.5em}
	\end{subfigure}

	\begin{subfigure}[t]{0.27\linewidth}
		\includegraphics[width=\textwidth]{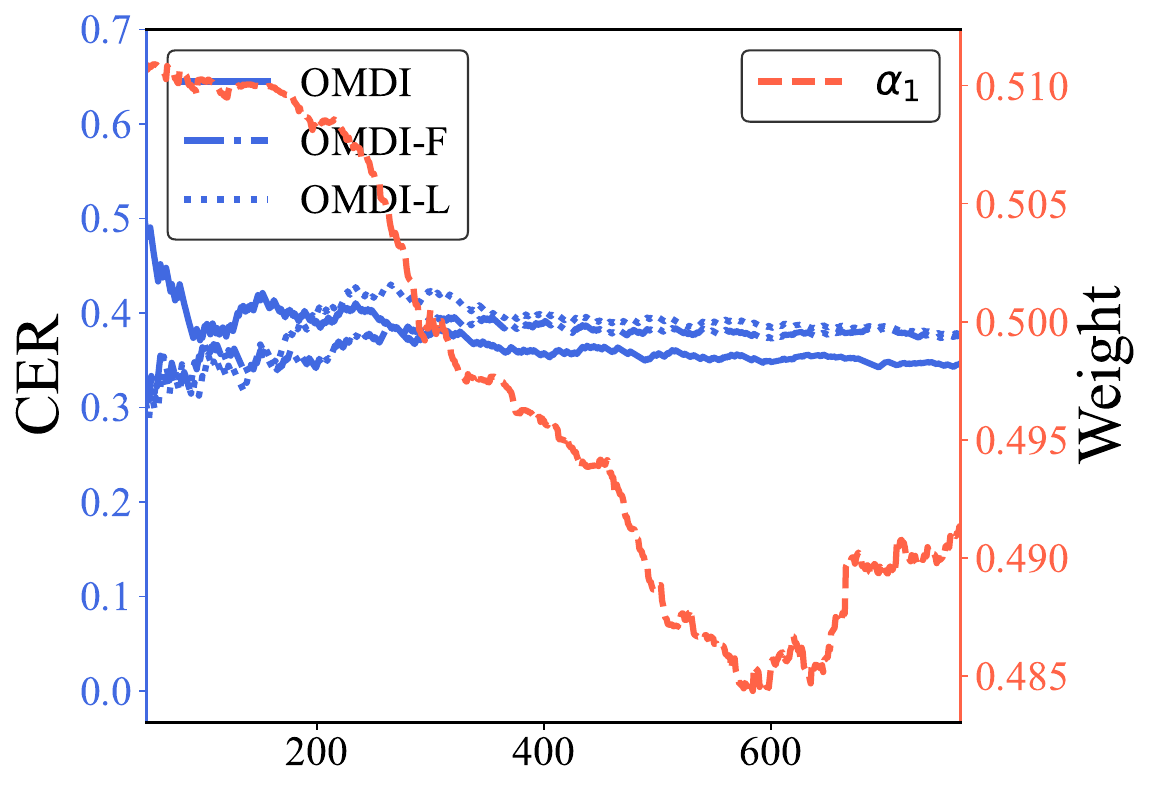}
		\caption{diabetes}
		\label{fig:Trap_diabetes_CER}
	\end{subfigure}
	\hspace{2em}
	\begin{subfigure}[t]{0.27\linewidth}
		\includegraphics[width=\textwidth]{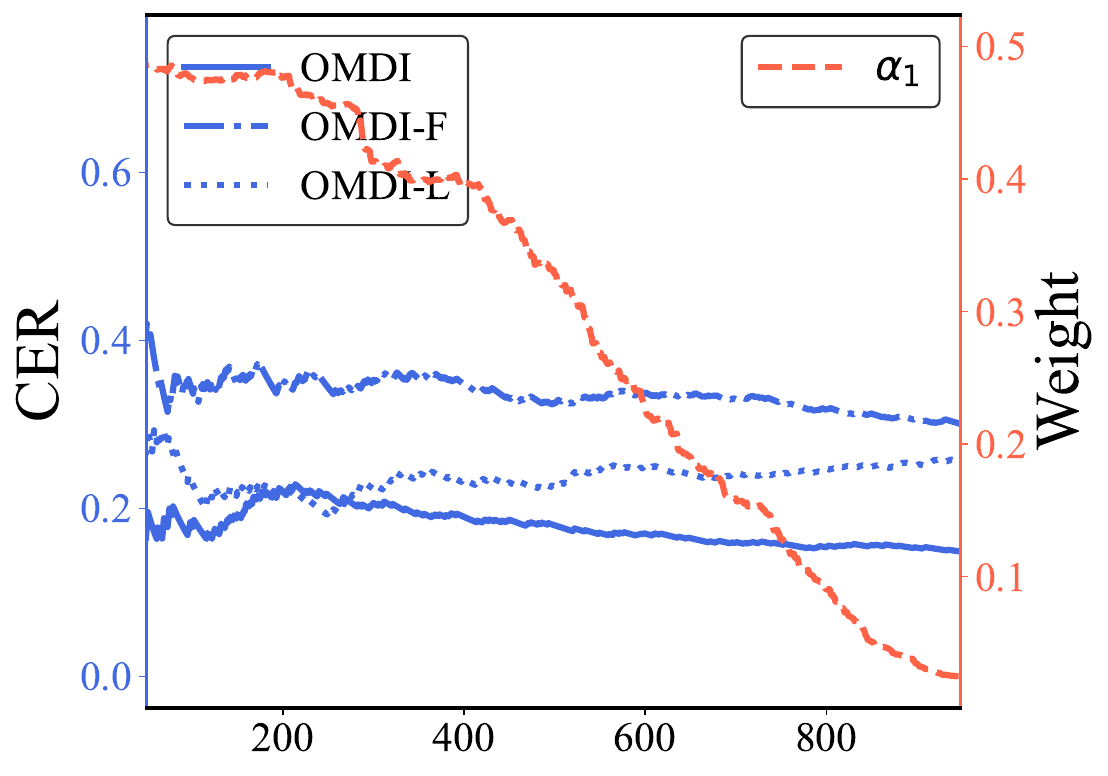}
		\caption{dna}
		\label{fig:Trap_dna_CER}
	\end{subfigure}
	\hspace{2em}
	\begin{subfigure}[t]{0.27\linewidth}
		\includegraphics[width=\textwidth]{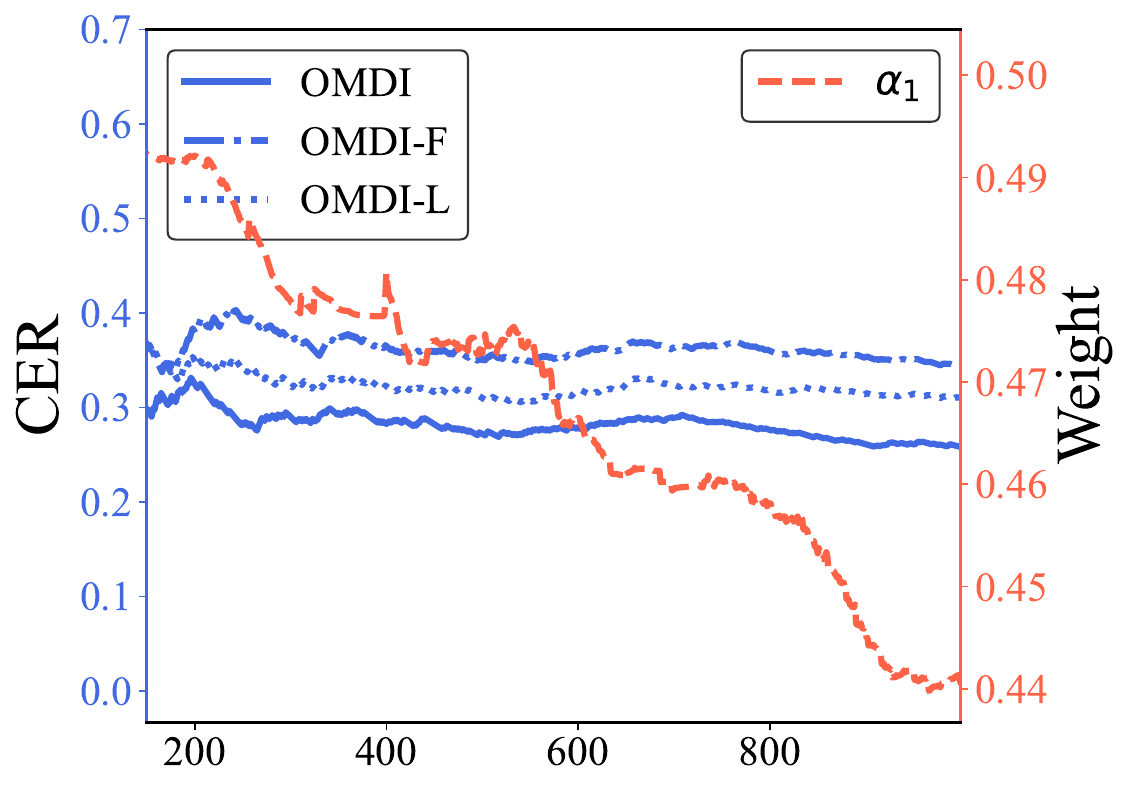}
		\caption{german}
		\label{fig:Trap_german_CER}
        \vspace{0.5em}
	\end{subfigure}

	\begin{subfigure}[t]{0.27\linewidth}
		\includegraphics[width=\textwidth]{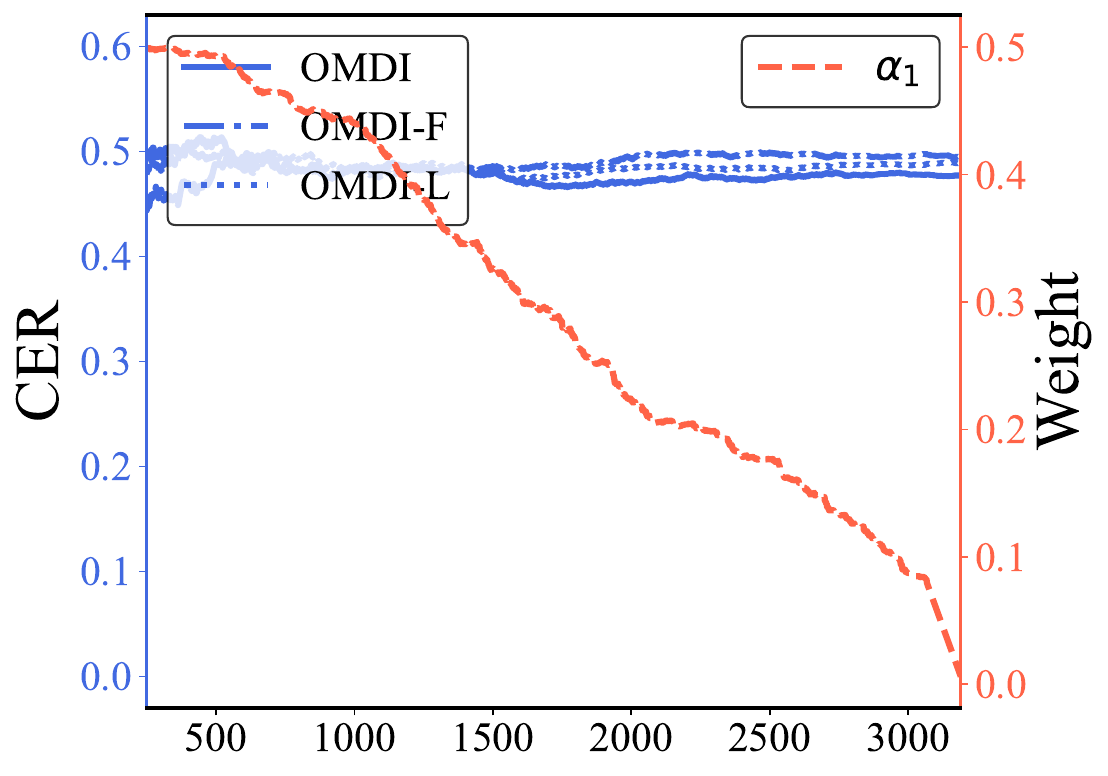}
		\caption{splice}
		\label{fig:Trap_splice_CER}
	\end{subfigure}
	\hspace{2em}
	\begin{subfigure}[t]{0.27\linewidth}
		\includegraphics[width=\textwidth]{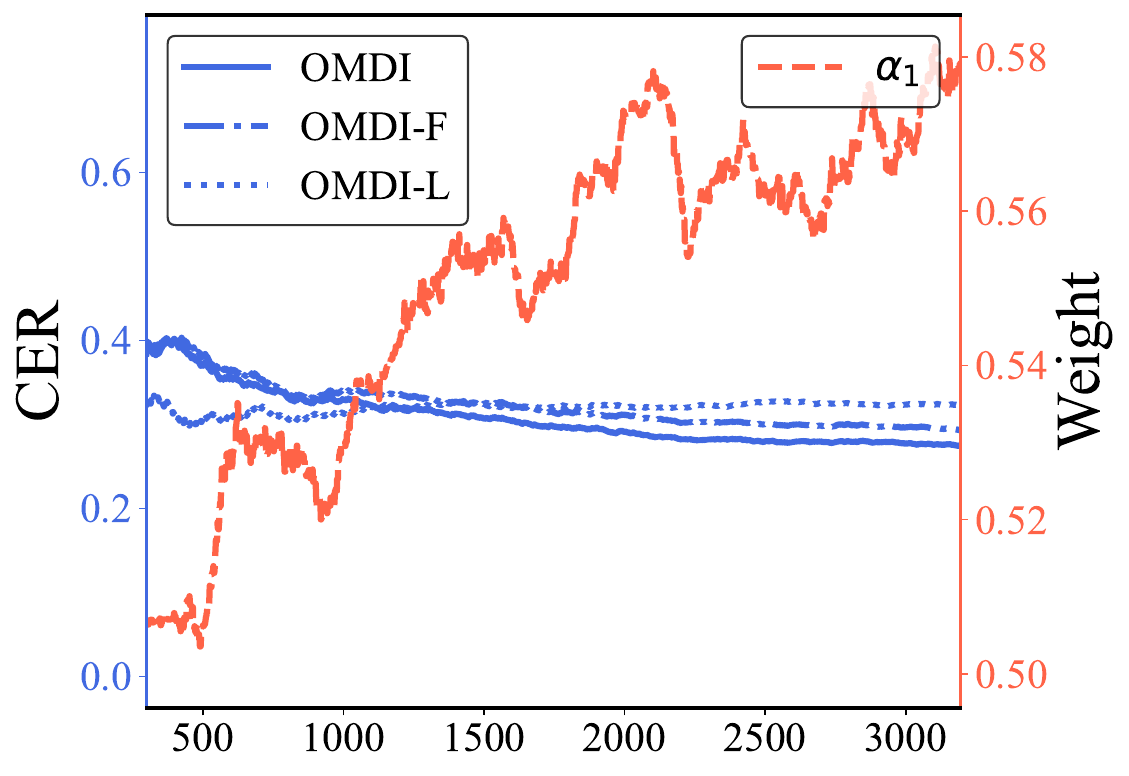}
		\caption{kr-vs-kp}
		\label{fig:Trap_kr-vs-kp_CER}
	\end{subfigure}
	\hspace{2em}
	\begin{subfigure}[t]{0.27\linewidth}
		\includegraphics[width=\textwidth]{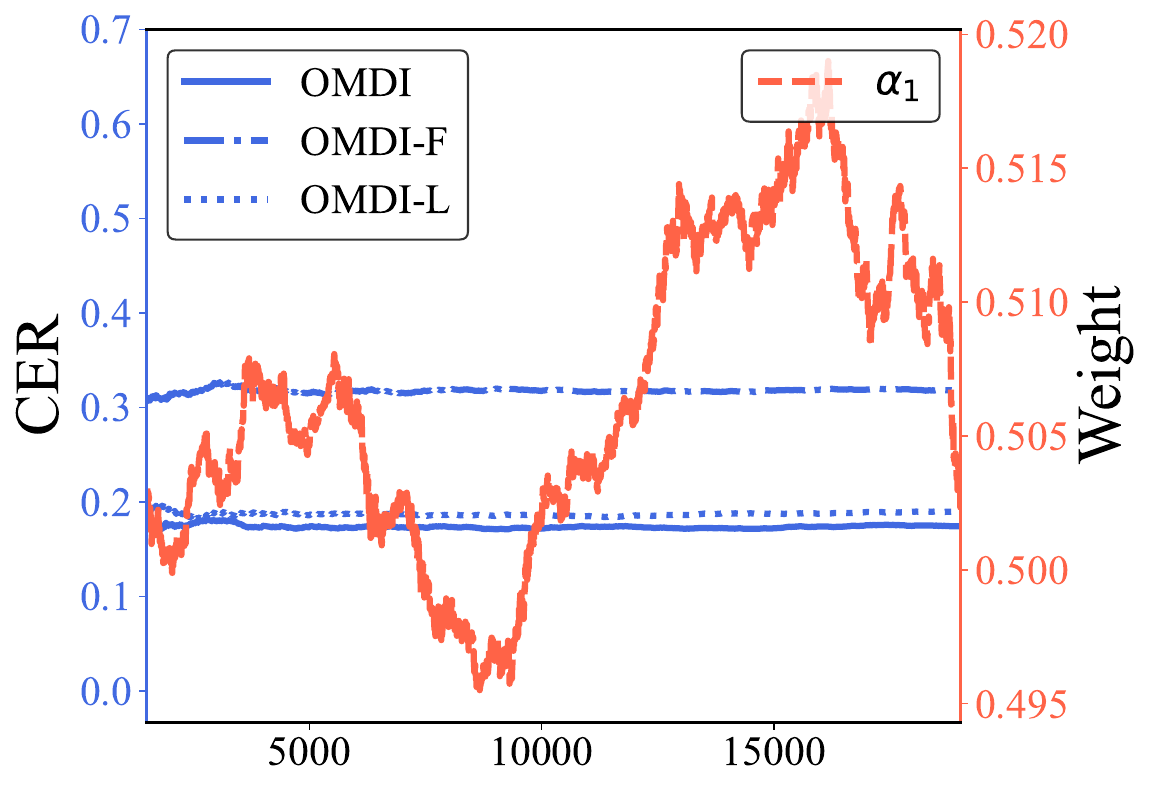}
		\caption{magic04}
		\label{fig:Trap_magic04_CER}
        \vspace{0.5em}
	\end{subfigure}

	\begin{subfigure}[t]{0.27\linewidth}
		\includegraphics[width=\textwidth]{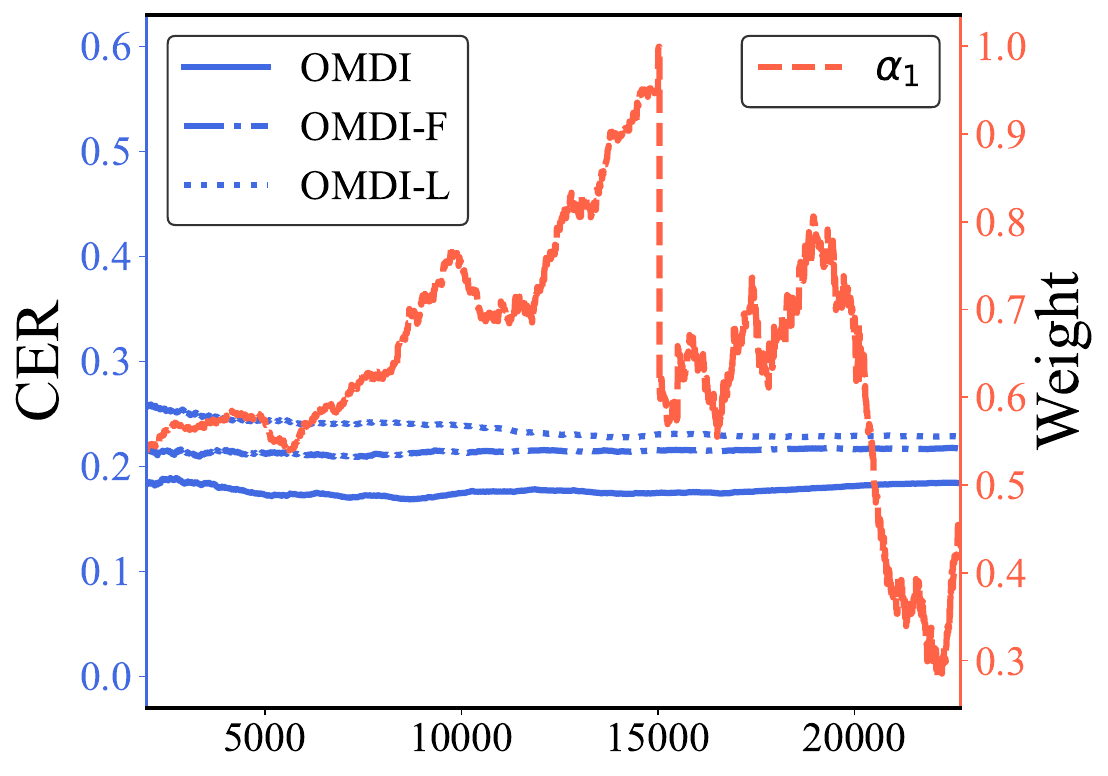}
		\caption{a8a}
		\label{fig:Trap_a8a_CER}
	\end{subfigure}
	\hspace{2em}
	\begin{subfigure}[t]{0.27\linewidth}
		\includegraphics[width=\textwidth]{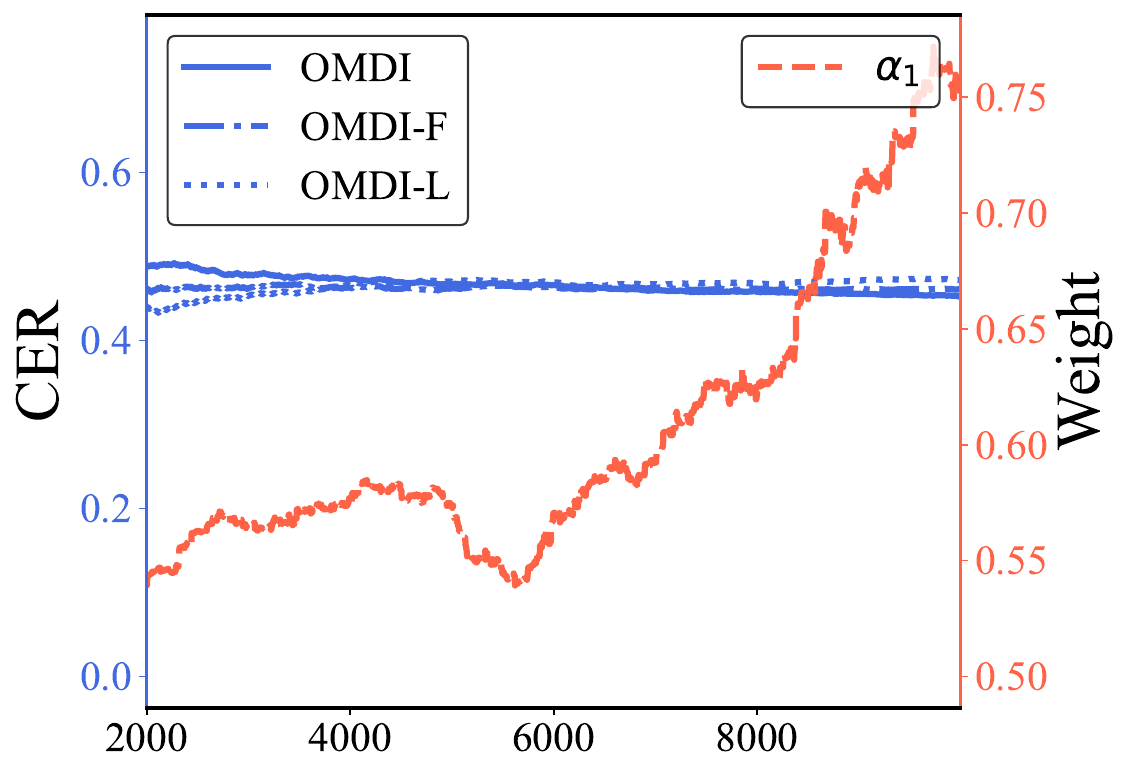}
		\caption{stream}
		\label{fig:Trap_stream_CER}
        \vspace{0.5em}
	\end{subfigure}
	
	\caption{
        Temporal variation of ensemble weight $\alpha_1$ and CERs of \alg~and its ablation variant \alg-F and \alg-L in all 14 trapezoidal data streams.
	}
	\label{fig:Trap_alpha_CER}
\end{figure*}

\begin{figure*}[!t]
	\centering
	\begin{subfigure}[t]{0.4\linewidth}
		\includegraphics[width=\textwidth]{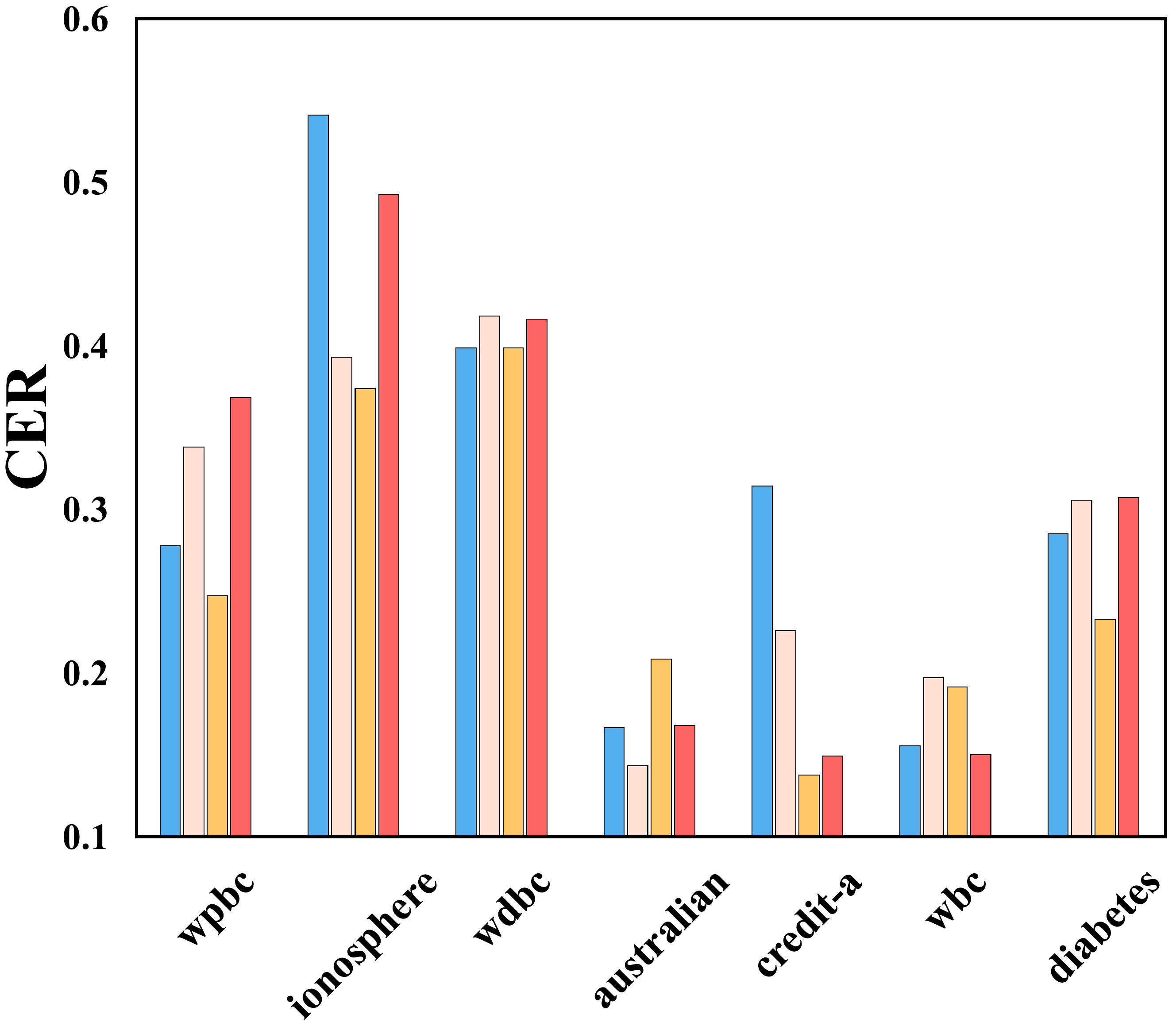}
		\label{fig:MissingCarp1}
	\end{subfigure}
	\begin{subfigure}[t]{0.41\linewidth}
		\includegraphics[width=\textwidth]{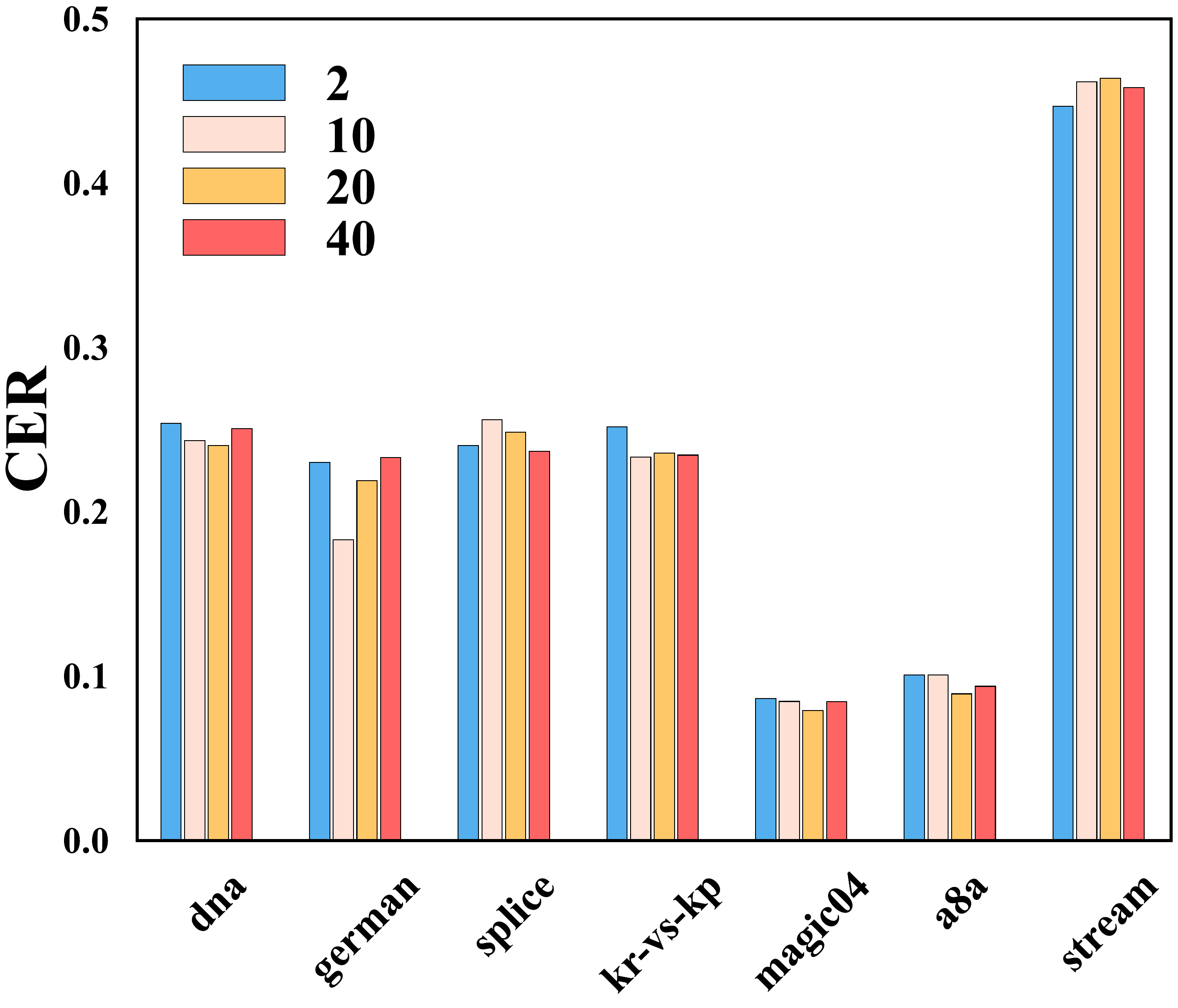}
		\label{fig:MissingCarp2}
	\end{subfigure}
	\caption{
    CER results of our algorithm (capricious data streams) in terms of missing ratio of labeled data on different datasets.		
	}
	\label{fig:buffersize}
\end{figure*}

\begin{figure*}[!t]
	\centering
	\begin{subfigure}[t]{0.4\linewidth}
		\includegraphics[width=\textwidth]{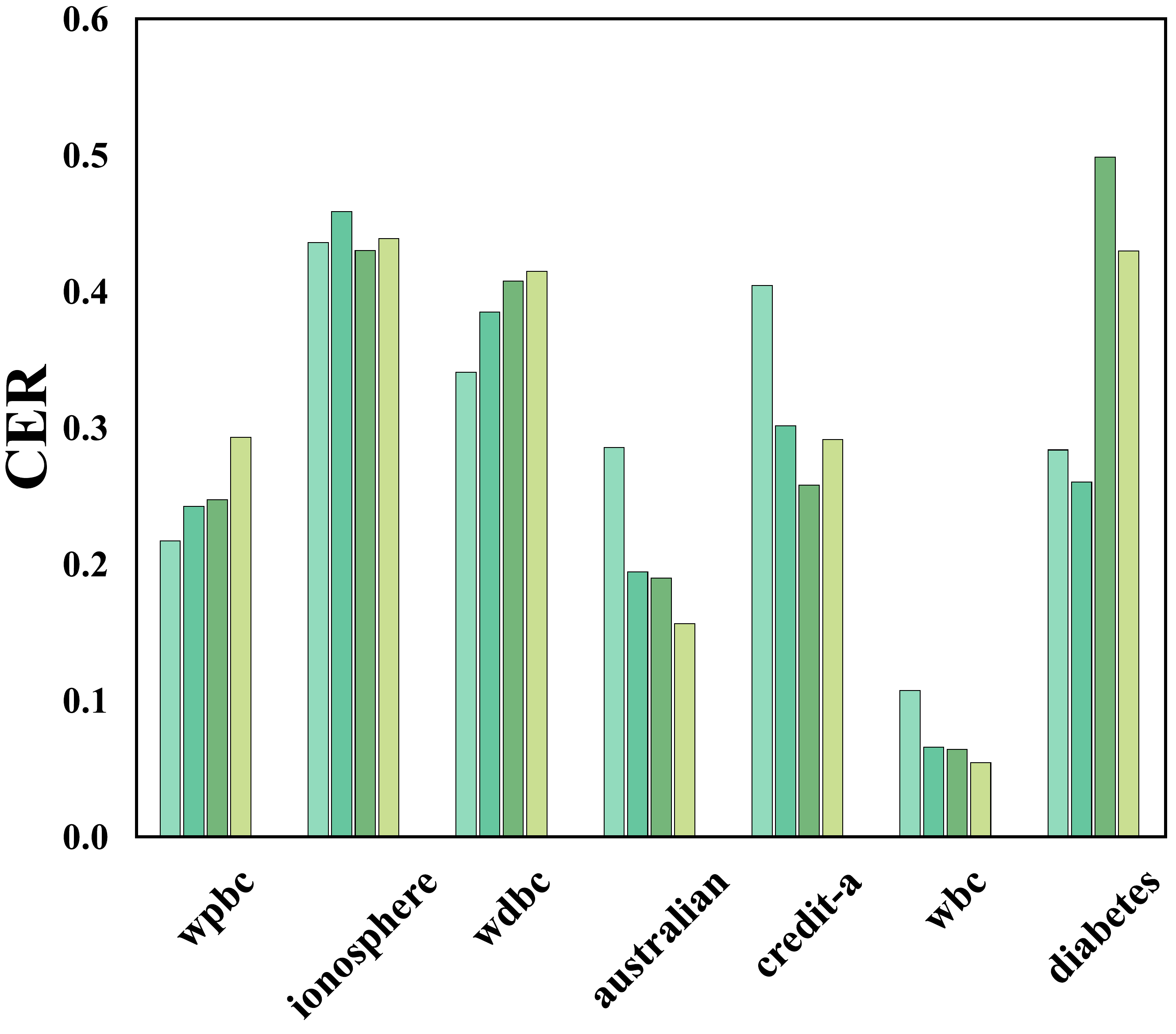}
		\label{fig:MissingCarp1}
	\end{subfigure}
	\begin{subfigure}[t]{0.41\linewidth}
		\includegraphics[width=\textwidth]{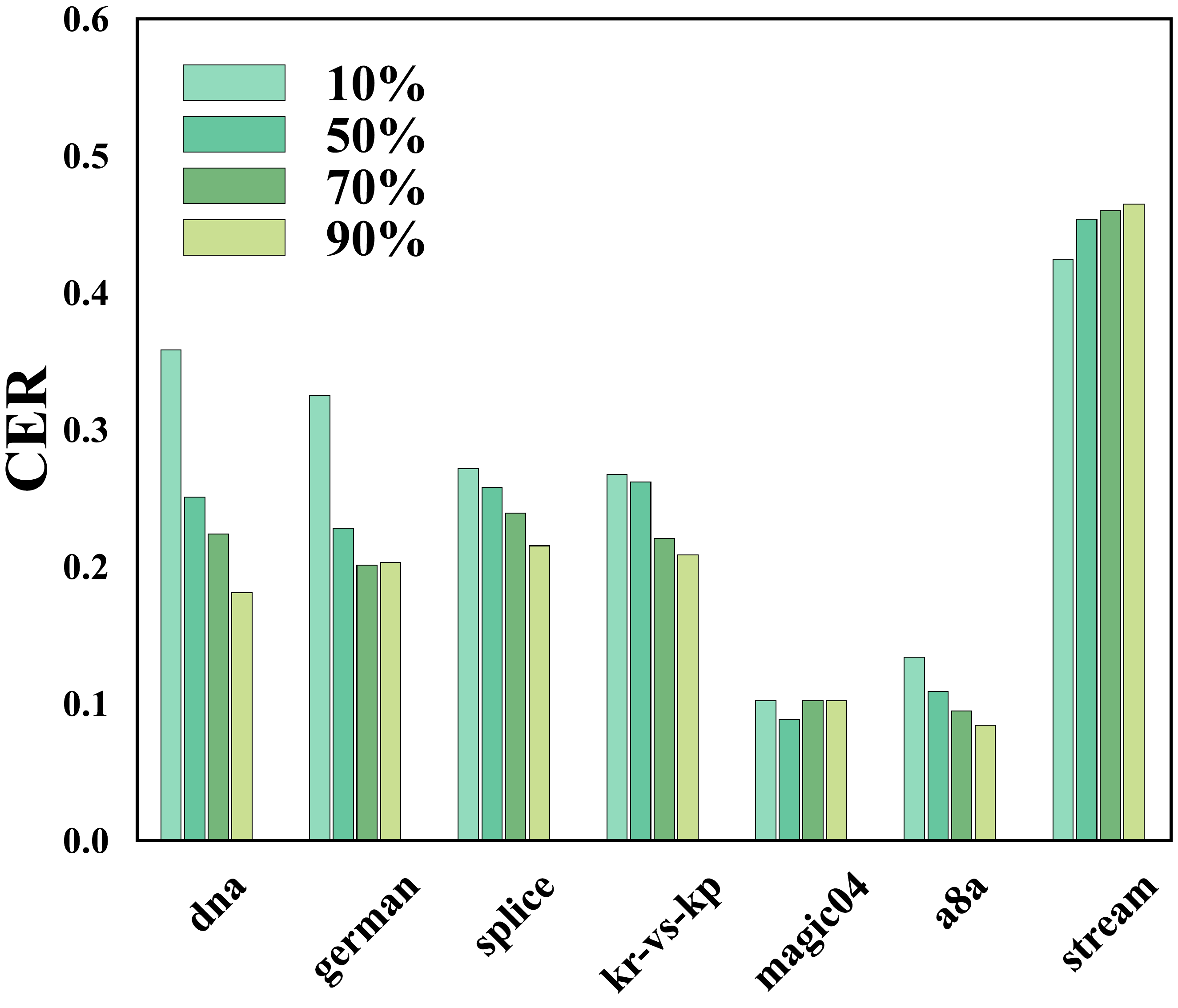}
		\label{fig:MissingCarp2}
	\end{subfigure}
	\caption{
		CER results of our algorithm (capricious data streams) in terms of missing ratio of labeled data on different datasets.
	}
	\label{fig:buffersize}
\end{figure*}

\begin{figure*}[!t]
	\centering
	\begin{subfigure}[t]{0.4\linewidth}
		\includegraphics[width=\textwidth]{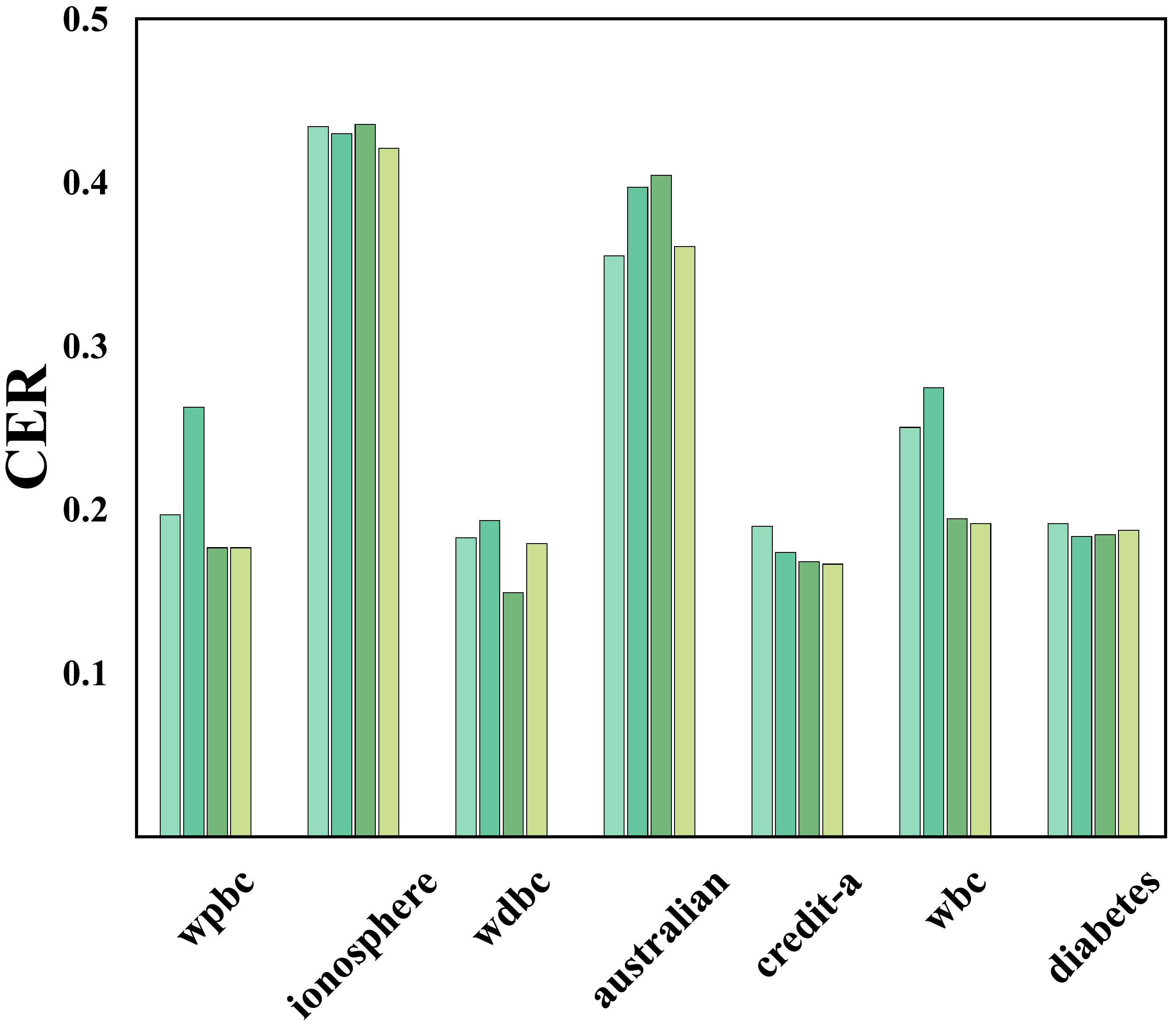}
		\label{fig:MissingCarp1}
	\end{subfigure}
	\begin{subfigure}[t]{0.41\linewidth}
		\includegraphics[width=\textwidth]{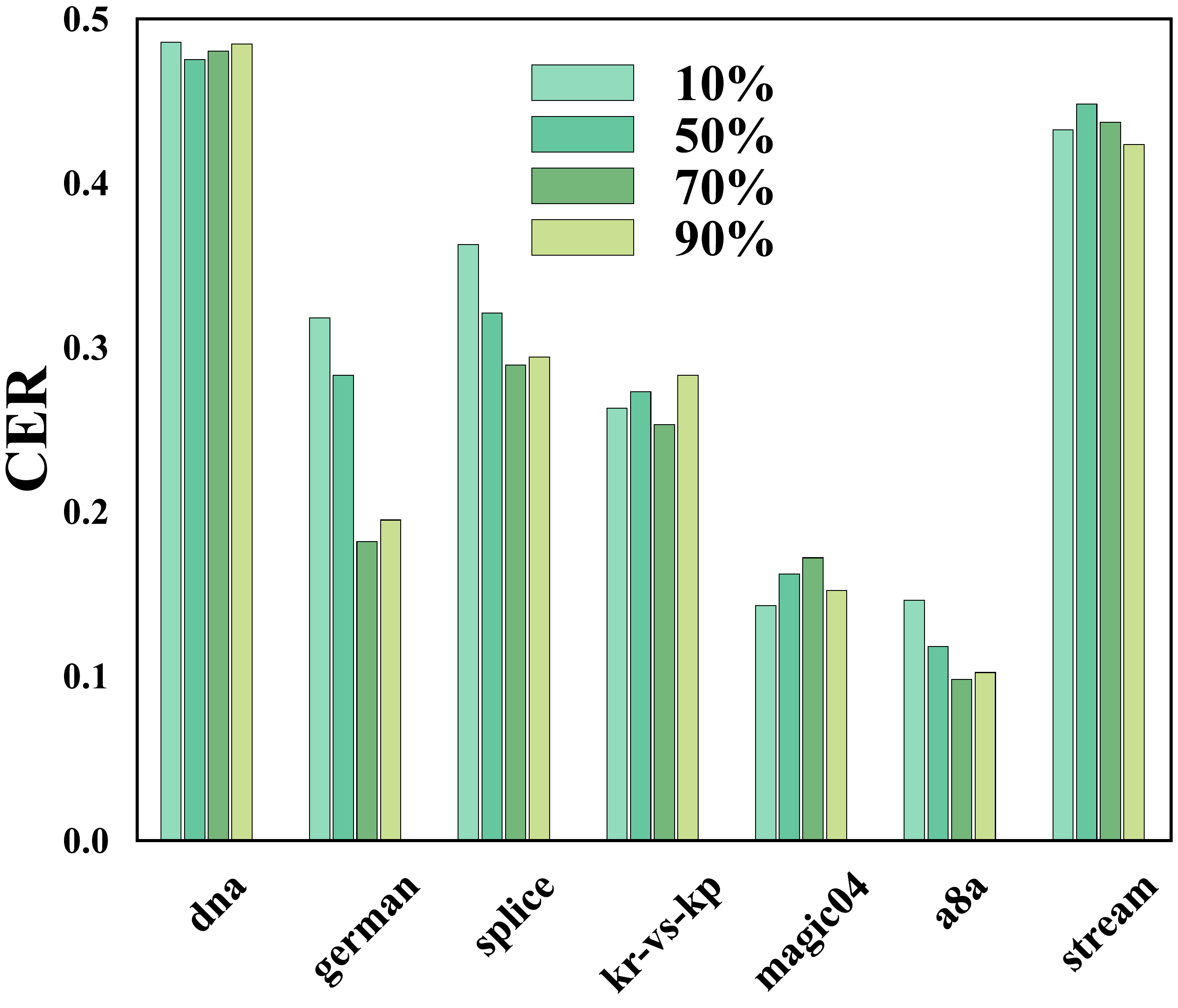}
		\label{fig:MissingCarp2}
	\end{subfigure}
	\caption{
		CER results of our algorithm (trapezoidal data streams) in terms of missing ratio of labeled data on different datasets.
	}
	\label{fig:buffersize}
\end{figure*}

\begin{table*}[!t]
	\centering
	\small
	\setlength{\tabcolsep}{5pt}
        \caption{
	CER results of our algorithm in terms of missing
	ratio of labeled data on different datasets.
	}
	\scalebox{.7}{
		\begin{tabular}{l|c|c|c|c|c|c|c|c|c|c|c|c|c}
	\toprule
	\midrule
		& &\multicolumn{6}{c}{Capricious Data Streams}  & \multicolumn {6}{|c}{Trapezoidal Data Streams}   \\
	\midrule
	Dataset                 & Missing & LR     & PAC    & Per    & Ridge  & SGDC   &\alg      & LR     & PAC    & Per    & Ridge  & SGDC   & \alg   \\ 
	\midrule
	\multirow{4}{*}{wpbc}   & 10\%    &.1919   &.3434   &.3181   &.2575   &.3636   &.2171     &.1617   &.2829   &.3484   &.2323   &.3484   &.1969   \\
	                        & 50\%    &.1262   &.2070   &.2474   &.2373   &.3131   &.2424     &.1011   &.3536   &.3637   &.2171   &.3637   &.2627   \\
	                        & 70\%    &.1363   &.2171   &.3737   &.1969   &.3333   &.2474     &.1364   &.2728   &.2627   &.2021   &.2627   &.1767   \\
	                        & 90\%    &.1212   &.2171   &.2828   &.1969   &.2121   &.2929     &.1162   &.2728   &.2930   &.1819   &.2930   &.1767   \\ 
	\midrule
	\multirow{4}{*}{ionosphere}  
	                        & 10\%    &.3675   &.3931   &.5384   &.2905   &.4188   &.4358     &.3521   &.3745   &.4891   &.3572   &.4583   &.4342   \\ 
							& 50\%    &.3276   &.3133   &.4701   &.4529   &.4188   &.4586     &.3734   &.3621   &.4821   &.3822   &.4512   &.4298   \\ 
							& 70\%    &.3961   &.3931   &.4501   &.4586   &.4103   &.4301     &.3625   &.3722   &.4682   &.3598   &.3821   &.4356   \\ 
							& 90\%    &.3761   &.3219   &.3048   &.4331   &.3789   &.4387     &.3356   &.3625   &.4529   &.3641   &.3921   &.4208   \\ 
	\midrule
	\multirow{4}{*}{wdbc}   & 10\%    &.3163   &.4112   &.3708   &.4481   &.4182   &.3409     &.1722   &.2618   &.3954   &.1827   &.3954   &.1827   \\ 
							& 50\%    &.3743   &.3866   &.3848   &.3268   &.4938   &.3848     &.1441   &.3145   &.3374   &.1265   &.3374   &.1933   \\ 
							& 70\%    &.4077   &.3989   &.4077   &.3760   &.3760   &.4077     &.1423   &.3691   &.3163   &.1054   &.3163   &.1493   \\ 
							& 90\%    &.4446   &.3954   &.4130   &.3866   &.4077   &.4147     &.1441   &.2917   &.3691   &.1142   &.3691   &.1792   \\ 
	\midrule
	\multirow{4}{*}{australia}  
							& 10\%    &.2869   &.2724   &.2826   &.2376   &.3144   &.2855     &.3072   &.3898   &.3478   &.2681   &.3478   &.3551   \\ 
							& 50\%    &.2057   &.1304   &.2884   &.1913   &.2768   &.1942     &.3232   &.4101   &.3898   &.3492   &.3898   &.3971   \\ 
							& 70\%    &.1507   &.1927   &.2840   &.1681   &.2318   &.1898     &.3464   &.4289   &.4478   &.3421   &.4478   &.4043   \\ 
							& 90\%    &.1202   &.2043   &.1565   &.1434   &.2188   &.1565     &.3667   &.4131   &.4188   &.3202   &.4188   &.3608   \\ 
	\midrule
	\multirow{4}{*}{credit-a}  
							& 10\%    &.1536   &.1971   &.3231   &.2594   &.2001   &.4043     &.2261   &.2478   &.3217   &.1753   &.3217   &.1898   \\ 
							& 50\%    &.0681   &.1565   &.2521   &.2333   &.1652   &.3014     &.2159   &.2594   &.3072   &.1841   &.3072   &.1739   \\ 
							& 70\%    &.0347   &.1753   &.1246   &.2608   &.3086   &.2579     &.2086   &.2841   &.3042   &.1637   &.3043   &.1681   \\ 
							& 90\%    &.0623   &.1826   &.1246   &.2391   &.2652   &.2913     &.2131   &.2681   &.2971   &.1347   &.2971   &.1667   \\ 
	\midrule
	\multirow{4}{*}{wbc}    & 10\%    &.0929   &.1301   &.1888   &.0701   &.2503   &.1072     &.1902	&.2503	&.4148	 &.1588	  &.4148   &.2503   \\ 
							& 50\%    &.0400   &.0844   &.1144   &.0672   &.2017   &.0658     &.2217	&.3175	&.3218	 &.1945	  &.3218   &.2746   \\ 
							& 70\%    &.0386   &.0801   &.0844   &.0501   &.0872   &.0643     &.2031	&.3276	&.3291	 &.1545	  &.3291   &.1945   \\ 
							& 90\%    &.0401   &.0815   &.2346   &.0401   &.1115   &.0543     &.2103	&.3204	&.3562	 &.1731	  &.3562   &.1914   \\ 
	\midrule
	\multirow{4}{*}{diabetes}  
							& 10\%    &.0807   &.3984   &.2877   &.2813   &.3033   &.2838     &.2382	&.3321	&.4192	&.1263	&.4192	&.1914   \\ 
							& 50\%    &.1576   &.3958   &.3294   &.0182   &.4701   &.2604     &.1315	&.3125	&.3359	&.1172	&.3359	&.1835   \\ 
							& 70\%    &.1119   &.4114   &.5104   &.0807   &.3033   &.4986     &.1406	&.3191	&.3294	&.151	&.3294	&.1848   \\ 
							& 90\%    &.1002   &.6171   &.3835   &.0143   &.0821   &.4296     &.1484	&.3216	&.3125	&.1458	&.3125	&.1875   \\ 
	\midrule
	\multirow{4}{*}{dna}  	& 10\%    &.3519   &.3898   &.3836   &.3391   &.3688   &.3582     &.3656	&.4394	&.4711	&.4415	&.4711	&.4857   \\ 
							& 50\%    &.2191   &.2528   &.2676   &.2455   &.2845   &.2507     &.4563	&.4668	&.4467	&.4752	&.4467	&.4752   \\ 
							& 70\%    &.2044   &.2107   &.2371   &.2044   &.2044   &.2339     &.4636	&.4836	&.492	&.4742	&.4921	&.4805   \\ 
							& 90\%    &.1717   &.1854   &.1844   &.2023   &.2002   &.1812     &.4752	&.4878	&.4783	&.5026	&.4783	&.4847   \\ 						
	\midrule
	\multirow{4}{*}{german} & 10\%    &.1590   &.2671   &.2900   &.2890   &.2901   &.3250     &.2020	&.3950	&.2590	&.2080	&.2590	&.3180   \\ 
							& 50\%    &.2160   &.3110   &.2080   &.2180   &.2790   &.2280     &.2040	&.4030	&.3300	&.1970	&.3300	&.2830   \\ 
							& 70\%    &.1160   &.2600   &.2980   &.1680   &.2440   &.2010     &.1750	&.3640	&.3890	&.1250	&.3890	&.1820   \\ 
							& 90\%    &.1380   &.2500   &.2380   &.0870   &.1070   &.2030     &.1580	&.3720	&.3610	&.0770	&.3610	&.1950   \\ 
	\midrule
	\multirow{4}{*}{splice} & 10\%    &.2282   &.2846   &.0949   &.1520   &.3576   &.2717     &.2852	&.3513	&.3725	&.3647	&.3725	&.3627   \\ 
							& 50\%    &.2119   &.2893   &.3166   &.2576   &.1884   &.2579     &.2819	&.3477	&.3595	&.3081	&.3595	&.3211   \\ 
							& 70\%    &.2119   &.1949   &.1931   &.0887   &.2112   &.2391     &.2637	&.3347	&.3389	&.278	&.3389	&.2894   \\ 
							& 90\%    &.1733   &.1579   &.1576   &.0332   &.0752   &.2153     &.2702	&.3422	&.3712	&.6727	&.3712	&.2943   \\ 
	\midrule
	\multirow{4}{*}{kr-vs-kp}  
							& 10\%    &.3041   &.2834   &.3003   &.3144   &.3588   &.2675     &.3372	&.2934	&.3129	&.3294	&.3741	&.2632   \\ 
							& 50\%    &.2381   &.3341   &.2841   &.2199   &.2221   &.2618     &.2983	&.3523	&.2984	&.2832	&.3638	&.2732   \\ 
							& 70\%    &.1705   &.3078   &.2647   &.2268   &.2515   &.2205     &.3294	&.293	&.2739	&.2632	&.3492	&.2532   \\ 
							& 90\%    &.2049   &.2897   &.2409   &.2133   &.3141   &.2086     &.2831	&.2894	&.2533	&.2536	&.3282	&.2832   \\ 
	\midrule
	\multirow{4}{*}{magic04}& 10\%    &.0641   &.3906   &.2088   &.0591   &.1911   &.1022     &.0792	&.3183	&.2112	&.0817	&.2123	&.1429   \\ 
							& 50\%    &.0641   &.1871   &.1241   &.0831   &.2750   &.0886     &.0717	&.2831	&.1627	&.0891	&.2632	&.1623   \\ 
							& 70\%    &.0641   &.2515   &.0854   &.0831   &.2598   &.1021     &.0732	&.2735	&.1292	&.0782	&.2372	&.1721   \\ 
							& 90\%    &.0641   &.0802   &.4608   &.1607   &.1224   &.1021     &.0791	&.1692	&.3529	&.1263	&.1982	&.1523   \\ 
	\midrule
	\multirow{4}{*}{a8a}    & 10\%    &.1007   &.1216   &.1786   &.0689   &.1455   &.1339     &.1102	&.1216	&.1652	&.0762	&.1526	&.1462   \\ 
							& 50\%    &.0754   &.1354   &.1532   &.0692   &.1395   &.1091     &.0823	&.1354	&.1622	&.0982	&.1423	&.1182   \\ 
							& 70\%    &.0735   &.1181   &.1831   &.0769   &.1360   &.0948     &.0862	&.1092	&.1831	&.0632	&.1292	&.0982   \\ 
							& 90\%    &.0787   &.1676   &.1641   &.0570   &.0968   &.0844     &.0883	&.1582	&.1641	&.0612	&.0953	&.1023   \\ 
	\midrule
	\multirow{4}{*}{stream} & 10\%    &.3783   &.4143   &.2159   &.4559   &.3615   &.4246     &.3892	&.4372	&.2312	&.4429	&.3529	&.4326   \\ 
							& 50\%    &.4441   &.4573   &.3792   &.4598   &.3522   &.4539     &.4623	&.4821	&.3082	&.4231	&.3688	&.4482   \\ 
							& 70\%    &.4575   &.4641   &.3541   &.4615   &.4326   &.4599     &.4521	&.4621	&.3427	&.4522	&.4352	&.4371   \\ 
							& 90\%    &.4427   &.4581   &.3339   &.4084   &.4623   &.4647     &.4231	&.4232	&.3392	&.4238	&.4523	&.4236   \\ 
	\midrule
	loss/win                & --      & $51/5$  & $24/32$ & $19/36$ & $43/13$ & $23/33$ & $161/119$        & $43/13$  & $9/47$ & $12/44$ & $41/15$ & $12/44$ & $117/163$ \\	
    \midrule
	\bottomrule
	\end{tabular}}
    \label{tab:1missing}
\end{table*}

\clearpage
\bibliographystyle{Reference-Format}
\bibliography{ref}


\end{document}